\documentclass[10pt,twocolumn,letterpaper]{article}

\usepackage{iccv}              

\definecolor{iccvblue}{rgb}{0.21,0.49,0.74}
\usepackage[pagebackref,breaklinks,colorlinks,allcolors=iccvblue]{hyperref}

\def\confName{ICCV}
\def\confYear{2025}

\usepackage[utf8]{inputenc} 
\usepackage[T1]{fontenc}    
\usepackage{hyperref}       
\usepackage{url}            
\usepackage{booktabs}       
\usepackage{amsfonts}       
\usepackage{nicefrac}       
\usepackage{microtype}      
\usepackage{xcolor}         
\usepackage{multirow}
\usepackage{makecell}
\usepackage{tabu}
\usepackage{fontawesome}
\usepackage{graphicx}
\usepackage{amsmath}
\usepackage{algorithmic}
\usepackage[ruled,norelsize,vlined]{algorithm2e}
\usepackage{bibunits}
\usepackage{colortbl}

\title{Synergistic Prompting for Robust Visual Recognition with Missing Modalities}

\author{Zhihui Zhang$^1$\quad  Luanyuan Dai$^3$\quad Qika Lin$^4$\quad Yunfeng Diao$^5$ \quad Guangyin Jin$^6$\\ 
Yufei Guo$^7$\quad Jing Zhang$^8$\quad Xiaoshuai Hao$^{2,}$\thanks{Corresponding author: Xiaoshuai Hao.}\\
\small$^1$Beijing Institute of Technology \quad
$^2$Beijing Academy of Artificial Intelligence \\
\small $^3$Nanjing University of Science and Technology \quad
\small$^4$National University of Singapore \\
\small $^5$Systems Laboratory of Anhui Province, Hefei University of Technology \quad
\small $^6$National Innovative Institute of Defense Technology\\
\small$^7$Intelligent Science \& Technology Academy of CASIC  \quad
$^8$School of Computer Science, Wuhan University \\
{\tt\small 3220231441@bit.edu.cn \quad xshao@baai.ac.cn}
}

\begin{document}
\maketitle
\begin{abstract}
Large-scale multi-modal models have demonstrated remarkable performance across various visual recognition tasks by leveraging extensive paired multi-modal training data.
However, in real-world applications, the presence of missing or incomplete modality inputs often leads to significant performance degradation.
Recent research has focused on prompt-based strategies to tackle this issue; however, existing methods are hindered by two major limitations: (1) static prompts lack the flexibility to adapt to varying missing-data conditions, and (2) basic prompt-tuning methods struggle to ensure reliable performance when critical modalities are missing.
To address these challenges, we propose a novel \textbf{Sy}nergistic \textbf{P}rompting (\textit{\textbf{SyP}}) framework for robust visual recognition with missing modalities. The proposed SyP introduces two key innovations: (I) a \textit{\textbf{Dynamic Adapter}}, which computes adaptive scaling factors to dynamically generate prompts, replacing static parameters for flexible multi-modal adaptation, and (II) a \textit{\textbf{Synergistic Prompting Strategy}}, which combines static and dynamic prompts to balance information across modalities, ensuring robust reasoning even when key modalities are missing.
The proposed SyP achieves significant performance improvements over existing approaches across three widely-used visual recognition datasets, demonstrating robustness under diverse missing rates and conditions. Extensive experiments and ablation studies validate its effectiveness in handling missing modalities, highlighting its superior adaptability and reliability.
\vspace{-2em}
\end{abstract}    
\section{Introduction}
\label{sec:intro}
Multi-modal learning enhances a model's performance and understanding in complex tasks by integrating information from different modalities. 
In recent years, multi-modal learning has made remarkable progress~\cite{zhao2024survey,hao2023mixgen} in various fields such as cross-modal retrieval~\cite{jiang2023cross,hao2023dual,hao2023uncertainty}, captioning~\cite{kim2024you,song2024emotional,chen2024sharegpt4v}, and visual question answering~\cite{liu2023cross,sima2024drivelm,li2024configure}, largely driven by large-scale paired datasets~\cite{kiela2020hateful, wang2015recipe} and powerful pre-trained multi-modal transformers~\cite{radford2021learning, liu2021swin, arnab2021vivit}. 
However, real-world multi-modal applications often face challenges with incomplete modality inputs due to factors such as sensor failures, privacy concerns, and data collection difficulties~\cite{jose2024enhancing, anagnostopoulos2024multimodal, khan2024exploring}. 
This issue becomes particularly severe when the missing modality contains critical information that cannot be compensated for by the remaining modalities, leading to significant performance degradation. As a result, incomplete modalities significantly impact the reliability, accuracy, and safety of multi-modal models in practical applications~\cite{guo2024mfhod,cheng2024retrieval}.

\begin{figure}[t]
    \centering    
{\includegraphics[width=1\columnwidth]{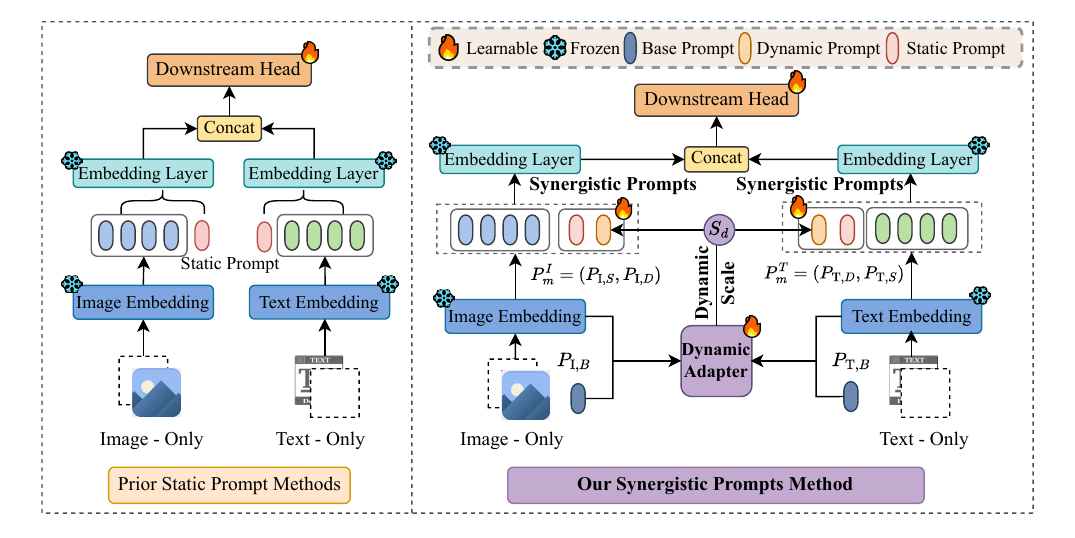}}  
    \vspace{-2em}
    \caption{\textbf{Comparison of Prior Static Prompt Methods and Our \textit{\textbf{SyP}} for Incomplete Multi-Modal Learning.} \textit{\textbf{SyP}} integrates static and dynamic prompts to balance information across modalities, ensuring robust reasoning even under missing key modalities.}
    \vspace{-1.5em}
    \label{fig:description}

\end{figure}

To address the missing-modality problem, researchers have explored three main categories of solutions: joint learning, cross-modal generation, and prompt-based methods. Joint learning approaches~\cite{zhao2021missing,kim2024missing} aim to distill cross-modal correlations by aligning latent feature spaces. However, these methods often rely on masking or imputation strategies, which introduce noise and reduce robustness~\cite{wu2024deep}. Similarly, cross-modal generation methods~\cite{le2024cross,meng2024multi} attempt to reconstruct missing modalities using available ones. Despite their potential, these methods struggle with modality heterogeneity---the inherent differences between modalities---resulting in poor reconstruction quality and high computational costs~\cite{zhang2024unified}.
Recently, prompt-based methods~\cite{lee2023multimodal, shi2025deep} have emerged as efficient and promising alternatives for addressing the challenging missing-modality problem. These methods leverage learnable prompts to adapt large pre-trained multi-modal transformers to missing-modality scenarios, requiring minimal additional parameters while preserving the rich knowledge encoded in the frozen backbone model~\cite{pipoli2025semantically,khattak2023maple,zhang2024dept}. However, existing prompt-based techniques face critical and significant limitations (see Fig.~\ref{fig:description}): (1) \textit{Static prompts are instance-agnostic}, applying the same prompt embeddings to all inputs regardless of the missing modality type, which severely limits adaptability to dynamic real-world conditions; and (2) \textit{simple prompt tuning lacks synergy between layers}, failing to fully exploit multi-modal dependencies across hierarchical model representations. 
These challenges strongly motivate the proposed SyP to develop a more robust and adaptive solution for effectively handling missing modalities.

In this paper, we propose \textit{\textbf{SyP~(Synergistic Prompting)}}, a novel framework to address limitations of static prompt-based methods in multi-modal learning under missing data conditions. \textit{\textbf{SyP}} introduces two key innovations: a dynamic adapter and a synergistic prompting strategy.
The dynamic adapter generates input-specific dynamic prompts by computing adaptive scaling factors based on available modality features, enabling the model to tailor responses to individual inputs and adapt effectively to diverse missing-modality scenarios. Unlike static prompts, which use fixed embeddings regardless of input variations, this dynamic approach ensures flexibility and robustness, maintaining performance even when critical modality data is absent. Complementing this, the synergistic prompting strategy combines static prompts, which provide a stable foundation of pre-trained knowledge, with dynamic prompts generated by the adapter, creating a dual-prompt mechanism that dynamically re-weights modality contributions.
This strategy ensures robust reasoning capabilities and accurate predictions under incomplete data conditions. The synergy between static and dynamic prompts enhances the model's resilience and enables seamless adaptation to real-world scenarios where missing data is prevalent. Extensive experiments and ablation studies across three widely-used visual recognition datasets demonstrate that the proposed \textit{\textbf{SyP}} achieves significant performance improvements over existing approaches, showcasing robustness under diverse missing rates and conditions, as well as superior adaptability and reliability in handling.

Our main contributions are summarized as follows:

\noindent $\bullet$ 
We propose \textit{\textbf{SyP (Synergistic Prompting)}}, a novel framework for robust visual recognition with missing modalities, combining static and dynamic prompts to ensure robust reasoning even with incomplete data.

\noindent $\bullet$  
Specifically, \textit{\textbf{SyP}} generates input-specific dynamic prompts via a dynamic adapter and combines them with static prompts through a synergistic strategy, enabling adaptive modality re-weighting for robust performance and maintained accuracy even with missing critical data.

\noindent $\bullet$ 
\textit{\textbf{SyP}} achieves significant performance gains across three datasets, demonstrating robustness under diverse missing rates and superior adaptability in handling missing modalities, introducing a new paradigm in multi-modal learning and paving the way for future research in the field.

\section{Related Work}
\label{sec:related_work}

\begin{figure*}[t]
  \centering
  \includegraphics[width=0.97\textwidth]{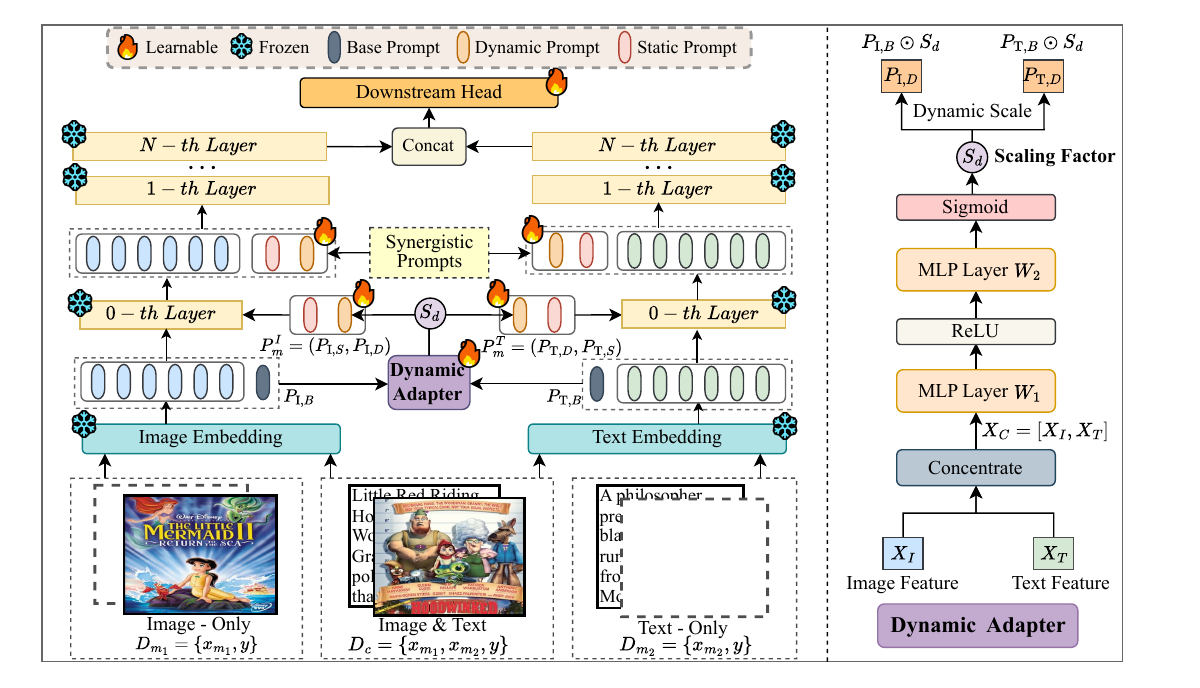}
\caption{\textbf{Overall Framework of Synergistic Prompting (SyP).}  
For each input, modality-specific prompts \( P_{m}^{I} \) and \( P_{m}^{T} \) are generated, where \( m \in \{c, m_1, m_2\} \) denotes the missing modality type. The prompts consist of dynamic prompts \( P_{\text{I},D}(P_{\text{T},D} \)), generated via a dynamic adapter, and static prompts \( P_{\text{I},S}(P_{\text{T},S} \)), integrated through a synergistic strategy. These prompts are prepended to input tokens \( x_{m_1} \) and \( x_{m_2} \), forming a unified sequence. 
The task-related tokens from both encoders are combined into a final representation and passed through a FC layer for prediction. During training, only the FC layer and synergistic prompts are updated, while the rest of the model remains frozen.
\vspace{-1em}
}
  \label{fig:model}
\end{figure*}

\textbf{Missing-Modality in Multi-modal Learning} 
Multi-modal learning has achieved significant progress~\cite{ji2025robobrain,li2024foundation,hao2023dual,lee2023multimodal,hao2023uncertainty,tan2025reason,hao2025mapfusion}, yet a critical challenge remains: handling incomplete modality inputs. This issue stems from sensor malfunctions, privacy concerns, or data collection difficulties~\cite{lang2025retrieval}. To tackle this challenge, researchers have developed methods broadly categorized into joint learning, cross-modal generation, and prompt-based approaches. 
Joint learning methods~\cite{kim2024missing,zhao2021missing} model inter-modal correlations to estimate missing modalities but often assume full modality availability during training, limiting their practicality~\cite{guo2024mfhod}. Cross-modal generation techniques~\cite{le2024cross,meng2024multi} regenerate missing modalities from available ones yet struggle with modality heterogeneity and computational overhead~\cite{zhang2024unified}. Prompt-based methods~\cite{pipoli2025semantically,shi2025deep}, which employ learnable prompts to guide pre-trained multi-modal transformers, have gained traction due to their efficiency. However, static prompts lack adaptability to the dynamic nature of missing modalities, often resulting in suboptimal performance when modalities are partially absent.
To address these issues, this work focuses on dynamic prompt adaptation and synergistic modality integration, aiming to overcome the limitations of static methods and improve robustness in missing-modality scenarios.

\textbf{Prompt Learning}
Prompt learning has emerged as a powerful technique for fine-tuning pre-trained models~\cite{sun2025strong,yang2025consistent,yao2024cpt,chen2024fine}, particularly in multi-modal learning. For example, CoOp~\cite{zhou2022learning} introduces learnable soft prompts alongside input images, enabling vision-language models to adapt to diverse vision tasks without retraining. In missing-modality scenarios, methods like MMP~\cite{lee2023multimodal}, MaPLe~\cite{khattak2023maple}, and DePT~\cite{zhang2024dept} utilize static prompts to guide pre-trained multi-modal transformers in handling incomplete data. However, static prompts are fixed and lack adaptability to the dynamic nature of missing data, limiting their effectiveness across diverse scenarios.
To address these limitations, dynamic prompt learning has gained traction~\cite{liu2025tackling,zhou2024dynamic}. Techniques such as DCP~\cite{shi2025deep} and RAGPT~\cite{lang2025retrieval} generate prompts based on input characteristics, enhancing model robustness in missing-modality scenarios. Building on these advancements, our approach introduces a synergistic prompting strategy that combines static and dynamic prompts, enabling the model to better manage missing modalities and achieve more robust performance in real-world applications. By leveraging the strengths of both static and dynamic approaches, our method ensures greater flexibility and adaptability, particularly in scenarios with variable modality availability.
\vspace{-1em}

\section{Methodology}
\label{sec:method}
\subsection{Preliminaries}

\textbf{Problem Definition}  
We address the missing-modality scenario in multi-modal learning. 
For simplicity and without loss of generality, we focus on multimodal inputs with \( M = 2 \) modalities, \( m_1 \) (\textit{e.g.}, image) and \( m_2 \) (\textit{e.g.}, text), though the framework can be extended to more modalities. 
Given a multimodal dataset \( D = \{ D_c, D_{m_1}, D_{m_2} \} \), we define \( D_c = \{ x_{m_1}, x_{m_2}, y \} \) as the modality-complete case, where \( x_{m_1} \) and \( x_{m_2} \) represent the features of the image and text modalities, respectively, and \( y \) is the label. 
The modality-incomplete cases are denoted as \( D_{m_1} = \{ x_{m_1}, y \} \) (missing text) and \( D_{m_2} = \{ x_{m_2}, y \} \) (missing image). 
The goal is to design a robust model capable of handling missing-modality scenarios, ensuring accurate predictions even when only one modality is available.

Fig.~\ref{fig:model} illustrates the framework of the proposed  SyP. We adopt the two-stream multimodal architecture CLIP~\cite{radford2021learning} as the backbone, ensuring generalizability. Input text and image are processed by CLIP's pretrained embedding layers, converting them into token sequences. 
For each input, modality-specific prompts \( P_{m}^{I} \) (image) and \( P_{m}^{T} \) (text)  are generated, where \( m \in \{c, m_1, m_2\} \) denotes the missing modality type. 
The prompts consist of \textit{dynamic prompts} \( P_{\text{I},D}(P_{\text{T},D} \)), generated via a dynamic adapter, and \textit{static prompts} \( P_{\text{I},S}(P_{\text{T},S} \)), integrated through a synergistic strategy. This enables adaptive modality re-weighting, ensuring robust performance even with missing data. The prompts are added to input tokens \( x_{m_1} \) and \( x_{m_2} \), forming a unified sequence. Task-related tokens from both encoders are combined into a final representation, passed through a fully-connected (FC) layer for prediction. Only the FC layer and synergistic prompts are updated during training, while the backbone remains frozen. 
The detailed architectural designs are described as follows.

\subsection{Dynamic Adapter}  
The dynamic adapter is a core component of the proposed SyP, designed to modulate base prompts by scaling them based on concatenated feature vectors from image and text encoders. The scaling factors are computed through a learned transformation of the combined modality features, enabling adaptive adjustment of the model's focus according to the strengths and weaknesses of each modality. This enhances the representation of relationships between different input types, improving overall performance.

\noindent \textbf{Feature Concatenation}  
Let \( X_{\text{I}} \in \mathbb{R}^{d_I} \) and \( X_{\text{T}} \in \mathbb{R}^{d_T} \) denote the image and text feature vectors, respectively. The concatenated feature vector \( X_{\text{C}} \) is formed by combining \( X_{\text{I}} \) and \( X_{\text{T}} \):  
\begin{equation}
    X_{\text{C}} = [X_{\text{I}}, X_{\text{T}}],
\end{equation}
where \( d_{\text{I}} \) and \( d_{\text{T}} \) are the dimensions of the image and text feature vectors. This combined vector \( X_{\text{C}} \) captures both visual and textual information, serving as input to the dynamic adapter for scaling factor computation.

\noindent \textbf{Scaling Factor Computation}  
The scaling factor \( S_d \) is computed using a multi-layer perceptron (MLP) with ReLU and sigmoid activation functions:  
\begin{equation}
S_d = \sigma \left( \text{MLP} \left( \text{ReLU} \left( \frac{1}{r} W_1 X_{\text{C}} + b_1 \right) \right) W_2 + b_2 \right),
\end{equation}
where \( W_1, W_2 \) are learned weights, \( b_1, b_2 \) are biases, and \( r \) is a \textit{reduction ratio}. The sigmoid function ensures \( S_d \in [0, 1] \), enabling adaptive modulation of prompt strength based on modality relevance.  

The scaling factor \( S_d \) determines the adaptive strength of the base prompts. A larger \( S_d \) enhances the prompt's influence, while a smaller \( S_d \) reduces it, allowing the model to dynamically adjust based on the relevance of each input.

\subsection{Synergistic Prompts}

Synergistic prompts consist of two primary components: \textit{dynamic prompts} and \textit{static prompts}. This combination enables the model to adaptively adjust the relative importance of information from different modalities, ensuring robust reasoning even when key modalities are sparse or absent.

\noindent \textbf{Dynamic Prompts} 
Dynamic prompts are central to the proposed SyP. They are generated by modulating base prompts for each modality using a \textit{dynamic adapter}. This allows the model to dynamically adjust its focus on image or text features based on learned scaling factors.
The base prompts, \( P_{\text{I},B} \) for images and \( P_{\text{T},B} \) for text, are modality-specific and learned during training. Initialized from image features and textual embeddings, respectively, they serve as foundational components for further refinement.

The scaling factors \( S_d \), computed by the dynamic adapter, are applied to the base prompts through element-wise multiplication:
\begin{equation}
P_{\text{I},D} = P_{\text{I},B} \odot S_d, \quad P_{\text{T},D} = P_{\text{T},B} \odot S_d,
\end{equation}
where \( \odot \) denotes element-wise multiplication.
This modulation adjusts the emphasis on prompts using a scaling factor, ensuring robustness when handling missing or sparse modalities. When a modality is absent, the scaling factor increases the corresponding prompt's weight, while in complete-modality scenarios, it reduces the weight, allowing the model to prioritize the features of available modalities.

\noindent \textbf{Static Prompts} 
The static prompt, denoted as \( P_S \), captures shared features between image and text modalities, facilitating cross-modal reasoning even when one modality is missing.
The static prompt \( P_S \) is projected into both image and text spaces using learned projection functions:
\begin{equation}
P_{\text{I}, S} = G_I(P_S), \quad P_{\text{T}, S} = G_T(P_S),
\end{equation}
where \( P_{\text{I}, S} \) and \( P_{\text{T}, S} \) are the projections into the image and text spaces, respectively. \( G_I \) and \( G_T \) are typically implemented as linear transformations, adapting the static prompt to each modality.

\noindent \textbf{Synergistic Prompts} 
The final synergistic prompts \( P_{m}^{I} \) (image) and \( P_{m}^{T} \) (text) are obtained by combining the dynamic prompts \( P_{\text{I},D} \) (\( P_{\text{T},D} \)) and static prompts \( P_{\text{I},S} \) (\( P_{\text{T},S} \)) through element-wise addition:
\begin{equation}
P_{m}^{I} = P_{\text{I},D} + P_{\text{I},S}, \quad P_{m}^{T} = P_{\text{T},D} + P_{\text{T},S}.
\end{equation}
This combination ensures that the model leverages both modality-specific adaptations (via dynamic prompts) and shared cross-modal features (via static prompts), enhancing its ability to handle diverse input scenarios, including missing or sparse modalities.

\subsection{Layer-wise Prompt Propagation}
Layer-wise prompt propagation creates a coherent information flow across transformer layers, enabling prompts to evolve at each level based on available input features. This ensures that each layer’s prompt integrates both current inputs and learned representations from previous layers, capturing hierarchical and nuanced features essential for multi-modal tasks involving visual and textual data.

\noindent \textbf{Propagation Mechanism} 
The propagation of prompts through layers is formalized for both image and text modalities. At the \( i \)-th layer (\( i \in [1, N] \)), the image prompt \( P_m^{\text{I}, R_i} \) is generated by applying a function \( F_i \) to the previous image prompt \( P_m^{\text{I}, R_{i-1}} \):
\begin{equation}
    P_m^{\text{I}, R_i} = F_i(P_m^{\text{I}, R_{i-1}}),
\end{equation}
where \( P_m^{\text{I}, R_i} \) is the prompt for the \( i \)-th layer of the image modality, \( F_i \) is a transformation function (e.g., linear projections and non-linear activations), and \( P_m^{\text{I}, R_{i-1}} \) is the prompt from the previous layer. Similarly, for the text modality, the recurrence relation is:
\begin{equation}
    P_m^{\text{T}, R_{i}} = G_i(P_m^{\text{T}, R_{i-1}}),
\end{equation}
where \( P_m^{\text{T}, R_{i}} \) is the prompt for the \( i \)-th layer of the text modality, \( G_i \) is the corresponding transformation function, and \( P_m^{\text{T}, R_{i-1}} \) is the prompt from the previous layer. This propagation ensures that prompts at each layer incorporate both previous prompts and newly extracted features, refining the model’s representation of the input across layers.

\noindent \textbf{Prompt Generation Functions} 
The functions \( F_i \) and \( G_i \) refine prompts through a series of operations. For the image modality, the generation function \( F_i \) is defined as:
\begin{equation}
    F_i(P_m^{\text{I}, R_{i-1}}) = \text{LN}( \text{FC}(\text{GeLU}(\text{FC}( P_m^{\text{I}, R_{i-1}} )))),
\end{equation}
where \( \text{FC} \) is a fully connected layer (bottleneck MLP), \( \text{GeLU} \) is the Gaussian Error Linear Unit activation function~\cite{hendrycks2016gaussian}, and \( \text{LN} \) is Layer Normalization~\cite{ba2016layer}. A similar function \( G_i \) is applied to the text modality, using its corresponding weights and biases. Through these layers, the model effectively captures both visual and textual information, enhancing its ability to handle multi-modal tasks.

\subsection{Training and Computational Efficiency}

The proposed SyP is computationally efficient, as it only updates the learnable prompts and the fully connected downstream layers during training. The parameters of the CLIP~\cite{radford2021learning} encoders (both image and text) remain frozen, significantly reducing computational costs compared to methods that fine-tune the entire model. The training process is guided by the total loss, defined as:
\begin{equation}
\mathcal{L}_{\text{total}} = \sum_{i=1}^{N} \mathcal{L}_i(P_{\text{final}}^{(i)}),
\end{equation}
where \( \mathcal{L}_i \) is the task-specific loss for the \( i \)-th sample, and \( P_{\text{final}}^{(i)} \) represents the final multi-modal prompt feature after processing by the prompt learner and transformer. This feature is computed as:
\begin{equation}
P_{\text{final}}^{(i)} = \text{FC} \left( P_m^{\text{I}, R_{i-1}} \mathop{\Vert} P_m^{\text{T}, R_{i-1}}   \right),
\end{equation}
where \( \Vert \) denotes the concatenation operation, \( P_m^{\text{I}, R_{i-1}} \) is the dynamic image prompt, and \( P_m^{\text{T}, R_{i-1}} \) is the dynamic text prompt for the \( i \)-th sample. Additionally, \( P_{\text{I}, S}^{(i)} \) and \( P_{\text{T}, S}^{(i)} \) represent the projected static prompts for the image and text modalities, respectively.

The concatenated vector of these prompts is passed through a fully connected layer (\( \text{FC} \)) to produce the representation \( P_{\text{final}}^{(i)} \) for each sample. These prompts are then added to the image and text features, forming a unified sequence that is passed through another fully connected (FC) layer for prediction. During training, only the FC layer and the synergistic prompts are updated, while the backbone remains frozen. This design ensures scalability, requiring fewer parameters while maintaining high performance.

\section{Experiment}
\label{sec:experiment}

\subsection{Experimental Details}

\noindent\textbf{Datasets} 
In this paper, we follow prior works~\cite{lee2023multimodal,shi2025deep} and evaluate our methods on three widely-used multi-modal datasets: MM-IMDb~\cite{arevalo2017gated}, UPMC Food-101~\cite{wang2015recipe}, and Hateful Memes~\cite{kiela2020hateful}. MM-IMDb provides 25,959 image-text annotated movies for multi-label genre classification, UPMC Food-101 offers noisy image-text pairs across 101 food categories for food classification, and Hateful Memes contains 10,000+ multi-modal examples for hateful meme detection, specifically designed to challenge uni-modal approaches. 
These datasets collectively enable comprehensive evaluation across diverse multi-modal tasks.

\textbf{Metrics} 
Following prior works~\cite{zhang2024dept,shi2025deep}, we adopt appropriate dataset-specific metrics to evaluate our method.
For MM-IMDb~\cite{arevalo2017gated}, we use F1-Macro to assess multi-label classification performance.
For UPMC Food-101~\cite{wang2015recipe}, we evaluate recognition performance using top-1 classification accuracy. 
For Hateful Memes~\cite{kiela2020hateful}, we employ the Area Under the Receiver Operating Characteristic Curve (AUROC).

\textbf{Implementation details} 
We adopt CLIP~\cite{radford2021learning} with a ViT-B/16~\cite{dosovitskiy2020image} image encoder as our multi-modal backbone. Images are resized to $224 \times 224$ and split into $16 \times 16$ patches, while text inputs are tokenized using CLIP's tokenizer with a maximum length of 77. Both image and text encoders are frozen, with only the prompts and fully connected layer tuned for the task. Image and text features ($d_{\text{I}}$ and $d_{\text{T}}$) have a dimension of 512, ensuring consistent representation across modalities. Learnable prompts of length $L_p = 36$ are added to features from $M=6$ layers, and missing modalities are replaced with zero-filled tensors. Training uses AdamW~\cite{loshchilov2017decoupled} with a learning rate of $1e^{-3}$, weight decay of $2e^{-2}$, and a 10\% warmup followed by linear decay over 20 epochs, with a batch size of 32 on four NVIDIA RTX 3090 GPUs.

\begin{table*}[t]
  \footnotesize	
  \setlength\tabcolsep{3pt}
  \centering
  \caption{Comparison with CoOp~\cite{zhou2022learning}, MMP~\cite{lee2023multimodal}, MaPLe~\cite{khattak2023maple}, DePT~\cite{zhang2024dept} and DCP~\cite{shi2025deep} on the MM-IMDb~\cite{arevalo2017gated}, UPMC Food-101~\cite{wang2015recipe}, and Hateful Memes~\cite{kiela2020hateful} datasets under various missing-modality cases with different missing rates. The bold number indicates the best performance. }
  \label{tab1}
  \begin{tabular}{c|c|cc|cccccc|cccccc}
      \hline 
      \multirow{2}{*}{ Datasets } & \multirow{2}{*}{\makecell[c]{  Missing \\ rate $\eta$}} & \multicolumn{2}{c|}{ Train/Test}  & \multicolumn{6}{c|}{ Validation set} & \multicolumn{6}{c}{ Testing set}\\
       & & Image & Text  & CoOp & MMP & MaPLe & DePT & DCP& \textbf{SyP~(Ours)} & CoOp & MMP & MaPLe & DePT&DCP& \textbf{SyP~(Ours)} \\
      \hline 
      \multirow{9}{*}{\makecell[c]{ 
      MM-IMDb \\(F1-Macro)}} & \multirow{3}{*}{$50 \%$} & $100 \%$ & $50 \%$  & 51.23 & 52.07 & 52.76 & 53.87 & 55.23 & \cellcolor{blue!5}{\textbf{56.16}} & 48.06 & 48.88 & 49.58 & 50.64 &52.13& \cellcolor{blue!5}{\textbf{53.90}} \\
      
      & & $50 \%$ & $100 \%$ & 53.04 & 54.52 & 55.26 & 56.04 & 57.32& \cellcolor{blue!5}{\textbf{58.55}} & 49.89 & 51.46 & 52.32 & 52.78 & 54.32& \cellcolor{blue!5}{\textbf{56.27}}\\
      
      & & $75 \%$ & $75 \%$ & 51.46 & 52.12 & 52.87 & 54.02 & 55.45& \cellcolor{blue!5}{\textbf{56.81}} & 48.37 & 49.32 & 49.56 & 50.87 & 52.32& \cellcolor{blue!5}{\textbf{55.02}}\\
      \cline{2-16}
      & \multirow{3}{*}{$70 \%$} &  $100 \%$ & $30 \%$ & 47.26& 48.23 &48.75 & 49.87 &51.35& \cellcolor{blue!5}{\textbf{53.34}} & 44.13 & 45.64 & 45.52 & 46.38 & 48.52& \cellcolor{blue!5}{\textbf{51.37}} \\

      & & $30 \%$ & $100 \%$ & 52.32 & 53.21 &53.98 & 55.04 & 56.21& \cellcolor{blue!5}{\textbf{56.89}} & 48.82 & 50.52 & 50.64 & 52.13 & 53.14 & \cellcolor{blue!5}{\textbf{54.20}}\\

      & & $65 \%$ & $65 \%$ & 50.22 & 51.34 & 52.31 & 53.17 & 54.24& \cellcolor{blue!5}{\textbf{55.22}} & 46.84 & 48.12 & 49.16 & 50.32& 51.42& \cellcolor{blue!5}{\textbf{52.90}} \\
      
      \cline{2-16}
      & \multirow{3}{*}{$90 \%$} & $100 \%$ & $10 \%$ & 47.86 & 48.84 &50.12 & 50.98 &52.36 & \cellcolor{blue!5}{\textbf{53.47}} & 44.76 & 45.32 & 46.84 & 47.56 & 49.26& \cellcolor{blue!5}{\textbf{50.21}}\\

      & & $10 \%$ & $100 \%$ & 51.65 & 52.36 &53.14 & 54.12 & 55.42& \cellcolor{blue!5}{\textbf{56.69}} & 48.32 & 49.12 & 50.13 & 50.88 & 52.22& \cellcolor{blue!5}{\textbf{53.72}}\\
      
      & & $55 \%$ & $55 \%$ & 47.44 & 48.04 &48.82 & 49.98 & 51.26& \cellcolor{blue!5}{\textbf{53.06}} & 44.12 & 44.87 & 45.12 & 46.54 & 48.04& \cellcolor{blue!5}{\textbf{49.63}} \\
      \hline 
      \multirow{9}{*}{\makecell[c]{Food101 \\(Accuracy)}} & \multirow{3}{*}{$50 \%$} & $100 \%$ & $50 \%$ & 77.36 & 78.24& 79.87 & 80.24 &  82.33&  \cellcolor{blue!5}{\textbf{83.47}} & 77.45 & 77.89 & 79.64 & 80.16 & 82.11& \cellcolor{blue!5}{\textbf{83.20}}  \\

      & & $50 \%$ & $100 \%$ & 86.98 & 87.12 & 87.48 & 87.85 & 89.23& \cellcolor{blue!5}{\textbf{89.81}} & 87.02 & 87.16 & 87.35 & 82.14 & 89.12& \cellcolor{blue!5}{\textbf{89.64}}\\

      & & $75 \%$ & $75 \%$ &  81.76 & 81.98 & 82.58 & 83.26 &85.25&\cellcolor{blue!5}{\textbf{86.33}} & 81.24 & 81.72 & 82.34 & 83.12 & 85.24& \cellcolor{blue!5}{\textbf{86.17}} \\
      
      \cline{2-16}
      & \multirow{3}{*}{$70 \%$} & $100 \%$ & $30 \%$ & 76.65 & 76.74 &76.87 & 76.87 & 79.18&\cellcolor{blue!5}{\textbf{80.09}} & 76.34 & 76.52 & 77.02 & 77.34 & 78.87& \cellcolor{blue!5}{\textbf{79.56}}\\

      & & $30 \%$ & $100 \%$ & 85.21 & 86.12 &86.36 & 86.52 &87.53&\cellcolor{blue!5}{\textbf{88.34}} & 84.78 & 85.64 & 85.89 & 86.12 & 87.32& \cellcolor{blue!5}{\textbf{88.67}}\\

      & & $65 \%$ & $65 \%$ & 79.14 & 79.56& 80.06 & 81.85 & 82.38& \cellcolor{blue!5}{\textbf{82.95}} & 78.87 & 79.12 & 79.84 & 81.46 & 81.87& \cellcolor{blue!5}{\textbf{82.45}} \\
      
      \cline{2-16}
      & \multirow{3}{*}{$90 \%$} & $100 \%$ & $10 \%$ & 72.65 & 73.74 &73.25 &74.22 & 75.54&\cellcolor{blue!5}{\textbf{76.46}} & 71.87 & 73.14 & 73.46 & 74.12& 75.26 & \cellcolor{blue!5}{\textbf{76.33}}\\

      & & $10 \%$ & $100 \%$ & 82.16 & 82.78 &83.42 & 84.02 &86.26&\cellcolor{blue!5}{\textbf{87.82}} & 81.67 & 82.14 & 83.12 & 83.56 & 85.78& \cellcolor{blue!5}{\textbf{86.41}} \\
      
      & & $55 \%$ & $55 \%$ & 77.36 & 77.78 &78.26 & 78.66 & 80.39& \cellcolor{blue!5}{\textbf{81.26}} & 76.46 & 76.58 & 77.85 & 78.12 & 79.87& \cellcolor{blue!5}{\textbf{81.03}} \\
      \hline 
      \multirow{9}{*}{\makecell[c]{Hateful \\ Memes  \\(AUROC)}} & \multirow{3}{*}{$50 \%$} & $100 \%$ & $50 \%$ & 58.32 & 58.56 & 58.78 & 59.31 & 60.24& \cellcolor{blue!5}{\textbf{66.41}}  & 60.56 & 60.31 &60.87 & 61.87 & 62.32& \cellcolor{blue!5}{\textbf{68.25}} \\

      & & $50 \%$ & $100 \%$ & 60.34 & 61.12 & 61.34 & 61.78 & 62.34& \cellcolor{blue!5}{\textbf{64.93}} & 62.41 & 62.35 & 63.13 & 63.88&64.46&\cellcolor{blue!5}{\textbf{66.80}}\\
      
      & & $75 \%$ & $75 \%$ & 62.34 & 62.87 & 63.14 & 63.24 & 63.78& \cellcolor{blue!5}{\textbf{68.09}} &64.87 & 65.84 &65.46 & 65.86 & 66.02& \cellcolor{blue!5}{\textbf{68.16}} \\
      \cline{2-16}
      & \multirow{3}{*}{$70 \%$} & $100 \%$ & $30 \%$ & 58.54 & 59.02 & 59.36 & 60.02 & 60.56&\cellcolor{blue!5}{\textbf{63.70}} & 60.74 & 61.12 &61.26 & 61.56 &62.82&\cellcolor{blue!5}{\textbf{68.94}} \\
      
      & & $30 \%$ & $100 \%$ & 60.12 & 60.78 & 61.32 & 61.54 & 62.32& \cellcolor{blue!5}{\textbf{66.79}} & 62.74 & 63.24 &63.14 & 63.48 &64.12&\cellcolor{blue!5}{\textbf{66.98}} \\
      & & $65 \%$ & $65 \%$ & 62.34 & 62.56 & 63.12 & 63.32 & 63.78& \cellcolor{blue!5}{\textbf{65.57}} & 64.82 & 65.04 &65.23 & 65.48 & 66.08& \cellcolor{blue!5}{\textbf{68.42}} \\
      \cline{2-16}
      & \multirow{3}{*}{$90 \%$} & $100 \%$ & $10 \%$ & 58.02 & 57.34 & 58.32 & 59.02 & 60.34& \cellcolor{blue!5}{\textbf{66.23}} & 60.03 & 57.21 &60.74 & 61.14 &62.08&\cellcolor{blue!5}{\textbf{69.70}} \\
      
      & & $10 \%$ & $100 \%$ & 59.02 & 59.32 & 60.21 & 60.56 & 61.34 & \cellcolor{blue!5}{\textbf{64.66}} & 61.46& 61.52 &61.87 & 62.42 &63.87&\cellcolor{blue!5}{\textbf{64.54}}\\
      
      & & $55 \%$ & $55 \%$ & 62.32 & 62.56 & 63.24 & 63.78 & 64.34& \cellcolor{blue!5}{\textbf{66.74}} & 64.32 & 63.34 &64.85 & 65.37 &66.78&\cellcolor{blue!5}{\textbf{68.93}} \\
      \hline
      \end{tabular}
      \label{tab:comparison}
     \vspace{-6px}
      \end{table*}

\textbf{Setating of Missing Pattern}
In this paper, we define the missing rate \( \eta \) as the proportion of modality-incomplete data relative to the entire dataset. This rate encompasses three potential cases of missing modalities: missing text, missing images, and missing both modalities. For cases of missing text or images, a missing rate of \( \eta \) indicates that \( \eta \) percent of the instances consist solely of text or images, with \( (1-\eta) \) percent containing both modalities. In the case of missing both modalities, \( \eta \) leads to \( \frac{\eta}{2} \) percent of instances being text-only, \( \frac{\eta}{2} \) percent being image-only, and \( (1-\eta) \) percent that are complete. This framework can be naturally extended to datasets with more modalities, using \( \frac{\eta}{M^2-2} \) for each missing case and \( (1-\eta) \) for complete data.

\textbf{Baselines}
We compare the proposed SyP with five competitive baselines: (1) CoOp~\cite{zhou2022learning}, which prepends prompts solely at the input level; (2) MMP~\cite{lee2023multimodal}, which inserts independent prompts for both input and intermediate features; (3) MaPLe~\cite{khattak2023maple}, which generates prompts in the image encoder based on those in the text encoder; (4) DePT~\cite{zhang2024dept}, which decouples base-specific knowledge from feature channels into an isolated space during prompt tuning; and (5) DCP~\cite{shi2025deep}, which enhances prompting through correlations, inter-layer relationships, and complementary semantics.

\subsection{Comparison with the State-of-the-arts}
To evaluate the effectiveness of the proposed SyP in various missing-modality scenarios, we conduct experiments on three widely used datasets: MM-IMDb~\cite{arevalo2017gated}, UPMC Food-101~\cite{wang2015recipe}, and Hateful Memes~\cite{kiela2020hateful}. We compare SyP with several state-of-the-art baselines, including CoOp~\cite{zhou2022learning}, MMP~\cite{lee2023multimodal}, MaPLe~\cite{khattak2023maple}, DePT~\cite{zhang2024dept} and DCP~\cite{shi2025deep}. We assess model performance across three missing rates ($\eta=50\%$, $\eta=70\%$, and $\eta=90\%$) and under three conditions: missing text, missing images, and both.

As shown in Tab.~\ref{tab1}, SyP consistently outperforms the state-of-the-art DCP~\cite{shi2025deep} across all missing conditions, particularly at high missing rates ($\eta=90\%$), 
showcasing superior dynamic prompt generation and synergistic prompt strategy, especially in scenarios where both modalities are missing.
Additionally, we observe dataset-specific differences in missing-modality impact: text missing leads to greater performance drops on MM-IMDB and Food101 due to their reliance on structured semantic cues, while in Hateful Memes, missing images result in sharper declines, highlighting the distinct roles of text and image modalities. This underscores the necessity for an adaptive prompt mechanism that dynamically adjusts to varying multi-modal settings.
Overall, our results demonstrate that SyP consistently outperforms all baselines across diverse missing-modality scenarios. The dynamic prompt adapter effectively captures critical information despite missing data, ensuring robust multi-modal understanding across datasets.

\begin{figure*}\small
    \centering
    \includegraphics[width=1\linewidth]{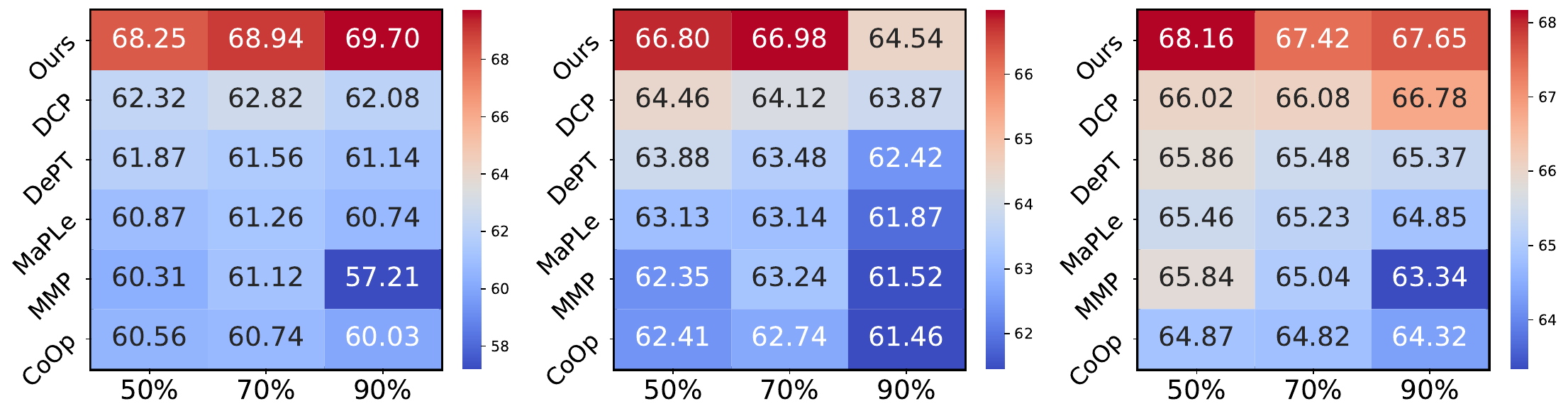}
    (a) Text Missing \hspace{3.5cm} (b) Image Missing \hspace{3.5cm} (c) Both Missing
    \caption{Generalization analysis on the Hateful Memes dataset across various missing rates in terms of AUROC}
    \label{fig:heatmap}
    \vspace{-1em}
\end{figure*}

\subsection{Ablation Study}
\noindent \textbf{Impact of Different Prompt Tuning} 
We explore the effectiveness of prompt tuning strategies under 50\% missing modality scenarios across three datasets (see Tab.~\ref{table:ablation1}). Results show that the baseline SyP (w/o Synergistic Prompts), relying on fixed encoders and fine-tuning only the classifier, performs poorly, highlighting the limitations of static methods. While Dynamic Prompts improve performance by adapting to missing modalities, Static Prompts offer only modest gains due to their rigidity. The SyP (w/ Synergistic Prompts) variant, combining dynamic and static prompts, achieves the best results by dynamically adjusting prompt weights, enabling robust multi-modal adaptation even with incomplete data. This study demonstrates the importance of integrating dynamic and static prompts for effective handling of missing modalities.

\noindent \textbf{Effectiveness of Dynamic Adapter}  
As shown in Tab.~\ref{table:ablation_adapter}, we conduct ablation experiments to evaluate the effectiveness of the dynamic adapter under 50\% missing modality scenarios. The results demonstrate three key findings:  
1) Using the \textit{base prompt} alone yields reasonable performance, but adding the \textit{dynamic adapter} improves results across all datasets, showcasing its ability to dynamically adjust prompts for missing modalities. 
2) The \textit{synergistic prompts} approach, which combines static and base prompts, further enhances performance, emphasizing the importance of balancing both prompt types.  
3) The \textit{dynamic adapter} integrated with synergistic prompts achieves the best results, confirming its critical role in enhancing the model’s adaptability to missing modalities. This combination enables robust multi-modal adaptation, maintaining strong comprehension and reasoning even under incomplete data conditions.

\begin{table}[t]
    \centering
    \setlength{\tabcolsep}{3pt}
    \resizebox{\linewidth}{!}{
    \begin{tabular}{lccc}
        \toprule
        \multirow{2}{*}{\textbf{Variant}} & {\textbf{Hateful Memes}} & {\textbf{Food101}}  & {\textbf{MM-IMDb}}\\ 
        & \textbf{(AUROC)} & \textbf{(Accuracy)} & \textbf{(F1\_Macro)} \\
         \midrule
        SyP~(w/o Synergistic Prompts) &57.35  & 71.59 &  44.63 \\
        SyP~(w/ only Dynamic Prompt) &  66.37   &   82.90 & 51.21  \\
        SyP~(w/ only Static Prompt)&  65.62 & 83.06   & 50.34 \\
        \rowcolor{blue!8} 
        \textbf{SyP~(w/ Synergistic Prompts)} & 	\textbf{68.16} & 	\textbf{86.17} &  \textbf{54.72}   \\
        \bottomrule
    \end{tabular}
    }
    \caption{Ablation study of prompt tuning under 50\% both missing on three datasets.}
    \label{table:ablation1}
    \vspace{-1em}
\end{table}

\begin{figure*}\small
    \centering
    \includegraphics[width=1\linewidth]{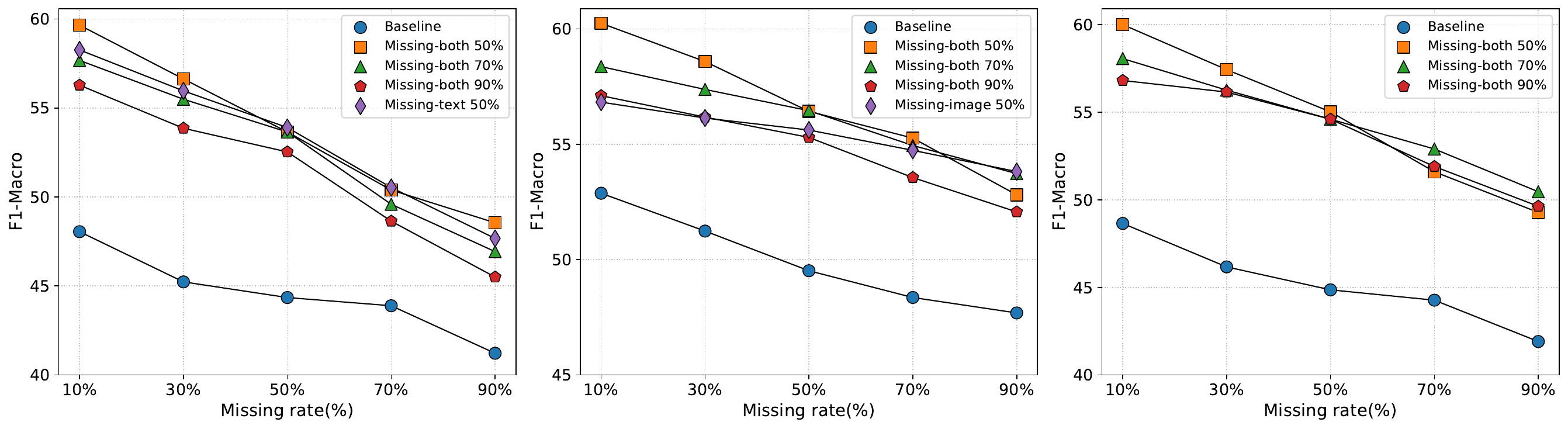}
    (a) Text Missing \hspace{3.5cm} (b) Image Missing \hspace{3.5cm} (c) Both Missing
    \vspace{-0.5em}
    \caption{\textbf{Generalizability Analysis of Our Method to Different Missing Rates on MM-IMDB dataset.} (a) Models are trained on missing-both or missing-text cases, and evaluated on missing-text cases with different missing rates. (b) Models are trained on missing-both or missing-image cases, and evaluated on missing-image cases with different missing rates. (c) All models are trained on missing-both cases, and evaluated on missing-both cases with different missing rates.}
    \vspace{-0.5em}
    \label{fig:line_chart}
\end{figure*}

\begin{figure*}\small
    \centering
    \includegraphics[width=1\linewidth]{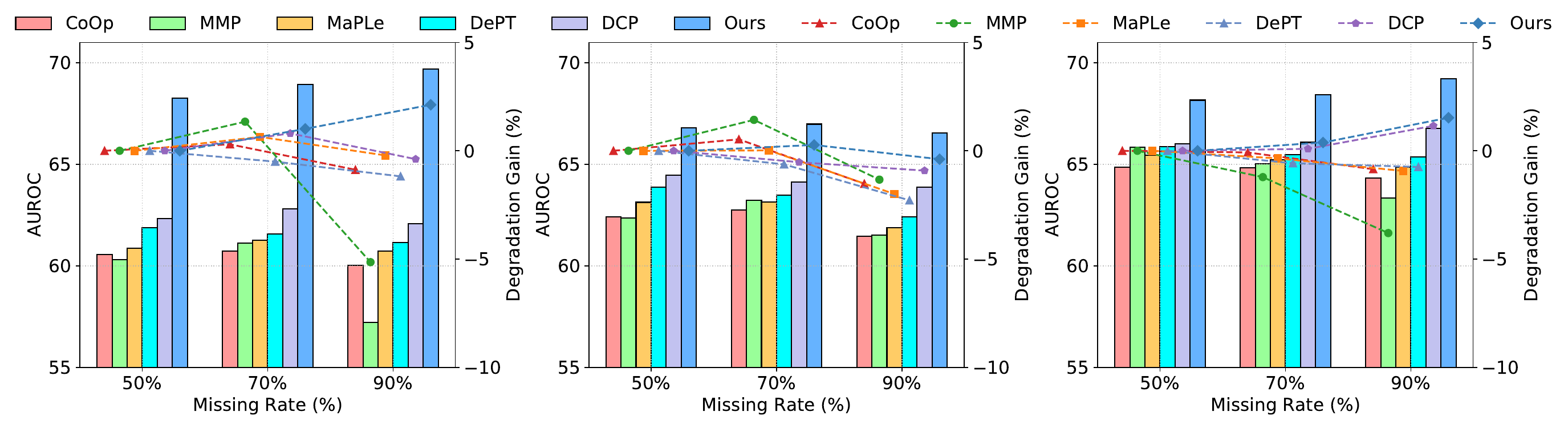}
    (a) Text Missing \hspace{3.5cm} (b) Image Missing \hspace{3.5cm} (c) Both Missing
    \vspace{-0.5em}
    \caption{Analysis of Robustness to Different Missing Rates compared with sate-of-the-art baselines on Hateful Memes dataset.}
    \vspace{-1.5em}
    \label{fig:Robustness}
\end{figure*}

 \noindent \textbf{Model Generalizability}
To investigate the generalizability of the proposed SyP, we conduct experiments with varying missing rates in the training set and evaluate its performance on a test set with matching missing rates. The results are visualized in a heatmap, covering three missing-modality conditions: text missing, image missing, and both missing.
In the text missing case (Fig.~\ref{fig:heatmap} (a) Text Missing), SyP consistently outperforms all baselines across missing rates of 50\%, 70\%, and 90\%, demonstrating the dynamic adapter's ability to handle missing text through dynamic prompt generation and information weight adjustment. For the image missing case (Fig.~\ref{fig:heatmap} (b) Image Missing), the proposed SyP achieves superior performance, though the gap narrows compared to text missing. In the both missing case (Fig.~\ref{fig:heatmap} (c) Both Missing), SyP maintains strong performance even when both modalities are absent, 
showcasing the robustness of the dynamic adapter and synergistic prompts strategy.
These results underscore the proposed SyP's effectiveness in handling missing data, maintaining consistent performance across varying missing rates and modalities. The dynamic prompts enables the model to identify critical cues and adjust to missing information, highlighting its generalizability. The consistent improvements over all baselines emphasize the proposed SyP's ability to sustain high understanding and reasoning capabilities despite significant input data loss.

\begin{table}[t]
    \centering
    \setlength{\tabcolsep}{2.5pt}
    \resizebox{\linewidth}{!}{
    \begin{tabular}{lccc}
        \toprule
        \multirow{2}{*}{\textbf{Variant}} & {\textbf{Hateful Memes}} & {\textbf{Food101}}  & {\textbf{MM-IMDb}}\\ 
        & \textbf{(AUROC)} & \textbf{(Accuracy)} & \textbf{(F1\_Macro)} \\
         \midrule
        SyP~(w/ only Base Prompt) &  64.27 & 81.68 & 48.95 \\
      \rowcolor{blue!8}   \textbf{SyP~(w/ only Base Prompt + Dynamic Adapter)}  &	\textbf{66.37}   &   \textbf{82.90} & \textbf{51.21} 	\\
        \midrule
        SyP~(w/ Synergistic Prompts)   & 66.19  & 	84.85   & 51.90 \\
        \rowcolor{blue!8} \textbf{SyP~(w/ Synergistic Prompts +  Dynamic Adapter)}& \textbf{68.16} & 	\textbf{86.17} &  \textbf{54.72} \\
        \bottomrule
    \end{tabular}
    }
    \caption{Ablation study of dynamic adapter under 50\% both missing on three datasets.}
    \label{table:ablation_adapter}
 \vspace{-1em}
\end{table}

 \noindent \textbf{Generalizability to Different Missing Rates} 
To validate the proposed SyP's generalizability to different missing rates, 
we conduct experiments on models trained under various missing-modality conditions and test across scenarios: text missing, image missing, and both missing on MM-IMDB dataset. The results, presented in Fig.~\ref{fig:line_chart}, assess performance across these scenarios.
Our findings highlight the effectiveness of proposed SyP. First, all model variants significantly outperform baselines across missing rates (10\%–90\%), demonstrating robustness in handling missing data. 
Models trained on a single missing modality perform more robustly under the same condition, indicating enhanced sensitivity to that modality's absence. Meanwhile, models trained under missing-both conditions exhibit strong performance across all scenarios, 
suggesting that synergistic prompts enable flexible adaptation, particularly when trained with higher missing rates.
In summary, proposed SyP demonstrates strong robustness when trained on missing-both cases, effectively handling diverse missing-modality situations. This underscores the value of our dynamic adapter approach, which adjusts information weights to ensure high performance across varying scenarios.

 \noindent \textbf{Robustness to Different Missing Rates} 
We conduct experiments to evaluate the proposed SyP's robustness against varying missing rates, comparing it with state-of-the-art baselines. As illustrated in Fig.~\ref{fig:Robustness}, all baselines exhibit significant performance degradation as the missing rate increases, underscoring their sensitivity to incomplete data. In contrast,  SyP demonstrates strong resilience, with only a slight decline in image-missing scenarios and even improved performance in text-missing and both-missing cases at higher missing rates. This highlights SyP's ability to effectively mitigate the impact of missing modalities, making it more reliable in real-world applications.
The robustness is attributed to the dynamic adapter mechanism, which generates dynamic prompts using adaptive scaling factors to dynamically adjust information weights across modalities. By balancing attention between modalities, SyP maintains robust reasoning capabilities even under missing data conditions. 
These results underscore the importance of dynamic adapter and synergistic prompts strategy, ensuring model's consistent performance across diverse missing-rate scenarios.

\section{Conclusion}
\label{sec:conclusion}

In this paper, we propose the Synergistic Prompting (SyP) framework to enhance multi-modal learning under missing-modality conditions. The proposed SyP introduces a Dynamic Adapter for adaptive prompt generation and a Synergistic Prompting Strategy to balance modality-specific information, ensuring robust reasoning with incomplete inputs. Extensive experiments on three visual recognition datasets show that the proposed SyP outperforms state-of-the-art methods, demonstrating superior robustness across diverse missing rates. Ablation studies confirm its effectiveness and adaptability, highlighting the importance of dynamic prompt generation and cross-modal information balancing. 
The framework’s ability to maintain performance under high missing rates makes it particularly suitable for real-world applications where data incompleteness is common. 
Future work will extend the framework to tasks with three or more modalities, further improving its scalability and applicability to complex multi-modal scenarios. 

\textbf{Acknowledgments}
This project has received funding from the NSF China (No. 62302139), and FRFCU China(No. PA2025IISL0113, JZ2025HGTB0227).

{
    \small
    \bibliographystyle{ieeenat_fullname}
    \bibliography{main}

\begin{thebibliography}{50}
\providecommand{\natexlab}[1]{#1}
\providecommand{\url}[1]{\texttt{#1}}
\expandafter\ifx\csname urlstyle\endcsname\relax
  \providecommand{\doi}[1]{doi: #1}\else
  \providecommand{\doi}{doi: \begingroup \urlstyle{rm}\Url}\fi

\bibitem[Anagnostopoulos et~al.(2024)Anagnostopoulos, Gkillas, Mavrokefalidis, Pikoulis, Piperigkos, and Lalos]{anagnostopoulos2024multimodal}
Christos Anagnostopoulos, Alexandros Gkillas, Christos Mavrokefalidis, Erion-Vasilis Pikoulis, Nikos Piperigkos, and Aris~S Lalos.
\newblock Multimodal federated learning in aiot systems: Existing solutions, applications, and challenges.
\newblock \emph{IEEE Access}, 2024.

\bibitem[Arevalo et~al.(2017)Arevalo, Solorio, Montes-y G{\'o}mez, and Gonz{\'a}lez]{arevalo2017gated}
John Arevalo, Thamar Solorio, Manuel Montes-y G{\'o}mez, and Fabio~A Gonz{\'a}lez.
\newblock Gated multimodal units for information fusion.
\newblock \emph{arXiv preprint arXiv:1702.01992}, 2017.

\bibitem[Arnab et~al.(2021)Arnab, Dehghani, Heigold, Sun, Lu{\v{c}}i{\'c}, and Schmid]{arnab2021vivit}
Anurag Arnab, Mostafa Dehghani, Georg Heigold, Chen Sun, Mario Lu{\v{c}}i{\'c}, and Cordelia Schmid.
\newblock Vivit: A video vision transformer.
\newblock In \emph{Proceedings of the IEEE/CVF international conference on computer vision}, pages 6836--6846, 2021.

\bibitem[Ba et~al.(2016)Ba, Kiros, and Hinton]{ba2016layer}
Jimmy~Lei Ba, Jamie~Ryan Kiros, and Geoffrey~E Hinton.
\newblock Layer normalization.
\newblock \emph{arXiv preprint arXiv:1607.06450}, 2016.

\bibitem[Chen et~al.(2024{\natexlab{a}})Chen, Li, Dong, Zhang, He, Wang, Zhao, and Lin]{chen2024sharegpt4v}
Lin Chen, Jinsong Li, Xiaoyi Dong, Pan Zhang, Conghui He, Jiaqi Wang, Feng Zhao, and Dahua Lin.
\newblock Sharegpt4v: Improving large multi-modal models with better captions.
\newblock In \emph{European Conference on Computer Vision}, pages 370--387, 2024{\natexlab{a}}.

\bibitem[Chen et~al.(2024{\natexlab{b}})Chen, Liu, Fournier-Viger, Zhang, Long, and Zhang]{chen2024fine}
Xiaojun Chen, Ting Liu, Philippe Fournier-Viger, Bowen Zhang, Guodong Long, and Qin Zhang.
\newblock A fine-grained self-adapting prompt learning approach for few-shot learning with pre-trained language models.
\newblock \emph{Knowledge-Based Systems}, 299:\penalty0 111968, 2024{\natexlab{b}}.

\bibitem[Cheng et~al.(2024)Cheng, Zhang, Xu, Trajcevski, Zhong, and Zhou]{cheng2024retrieval}
Zhangtao Cheng, Jienan Zhang, Xovee Xu, Goce Trajcevski, Ting Zhong, and Fan Zhou.
\newblock Retrieval-augmented hypergraph for multimodal social media popularity prediction.
\newblock In \emph{Proceedings of the 30th ACM SIGKDD conference on knowledge discovery and data mining}, pages 445--455, 2024.

\bibitem[Dosovitskiy et~al.(2020)Dosovitskiy, Beyer, Kolesnikov, Weissenborn, et~al.]{dosovitskiy2020image}
Alexey Dosovitskiy, Lucas Beyer, Alexander Kolesnikov, Weissenborn, et~al.
\newblock An image is worth 16x16 words: Transformers for image recognition at scale.
\newblock \emph{arXiv preprint arXiv:2010.11929}, 2020.

\bibitem[Guo et~al.(2024)Guo, Zhan, Jiang, Ge, Chen, Xu, Li, and Liu]{guo2024mfhod}
Jinxin Guo, Weida Zhan, Yichun Jiang, Wei Ge, Yu Chen, Xiaoyu Xu, Jin Li, and Yanyan Liu.
\newblock Mfhod: Multi-modal image fusion method based on the higher-order degradation model.
\newblock \emph{Expert Systems with Applications}, 249:\penalty0 123731, 2024.

\bibitem[Hao and Zhang(2023)]{hao2023uncertainty}
Xiaoshuai Hao and Wanqian Zhang.
\newblock Uncertainty-aware alignment network for cross-domain video-text retrieval.
\newblock \emph{Advances in Neural Information Processing Systems}, 36:\penalty0 38284--38296, 2023.

\bibitem[Hao et~al.(2023{\natexlab{a}})Hao, Zhang, Wu, Zhu, and Li]{hao2023dual}
Xiaoshuai Hao, Wanqian Zhang, Dayan Wu, Fei Zhu, and Bo Li.
\newblock Dual alignment unsupervised domain adaptation for video-text retrieval.
\newblock In \emph{Proceedings of the IEEE/CVF conference on computer vision and pattern recognition}, pages 18962--18972, 2023{\natexlab{a}}.

\bibitem[Hao et~al.(2023{\natexlab{b}})Hao, Zhu, Appalaraju, Zhang, Zhang, Li, and Li]{hao2023mixgen}
Xiaoshuai Hao, Yi Zhu, Srikar Appalaraju, Aston Zhang, Wanqian Zhang, Bo Li, and Mu Li.
\newblock Mixgen: A new multi-modal data augmentation.
\newblock In \emph{Proceedings of the IEEE/CVF winter conference on applications of computer vision}, pages 379--389, 2023{\natexlab{b}}.

\bibitem[Hao et~al.(2025)Hao, Diao, Wei, Yang, Hao, Yin, Zhang, Li, Zhao, and Liu]{hao2025mapfusion}
Xiaoshuai Hao, Yunfeng Diao, Mengchuan Wei, Yifan Yang, Peng Hao, Rong Yin, Hui Zhang, Weiming Li, Shu Zhao, and Yu Liu.
\newblock Mapfusion: A novel bev feature fusion network for multi-modal map construction.
\newblock \emph{Information Fusion}, 119:\penalty0 103018, 2025.

\bibitem[Hendrycks and Gimpel(2016)]{hendrycks2016gaussian}
Dan Hendrycks and Kevin Gimpel.
\newblock Gaussian error linear units (gelus).
\newblock \emph{arXiv preprint arXiv:1606.08415}, 2016.

\bibitem[Ji et~al.(2025)Ji, Tan, Shi, Hao, Zhang, Zhang, Wang, Zhao, Mu, An, et~al.]{ji2025robobrain}
Yuheng Ji, Huajie Tan, Jiayu Shi, Xiaoshuai Hao, Yuan Zhang, Hengyuan Zhang, Pengwei Wang, Mengdi Zhao, Yao Mu, Pengju An, et~al.
\newblock Robobrain: A unified brain model for robotic manipulation from abstract to concrete.
\newblock In \emph{Proceedings of the Computer Vision and Pattern Recognition Conference}, pages 1724--1734, 2025.

\bibitem[Jiang and Ye(2023)]{jiang2023cross}
Ding Jiang and Mang Ye.
\newblock Cross-modal implicit relation reasoning and aligning for text-to-image person retrieval.
\newblock In \emph{Proceedings of the IEEE/CVF Conference on Computer Vision and Pattern Recognition}, pages 2787--2797, 2023.

\bibitem[Jose et~al.(2024)Jose, Nguyen, and Medjaher]{jose2024enhancing}
Sagar Jose, Khanh~TP Nguyen, and Kamal Medjaher.
\newblock Enhancing industrial prognostic accuracy in noisy and missing data context: Assessing multimodal learning performance.
\newblock \emph{Journal of Intelligent Manufacturing}, pages 1--25, 2024.

\bibitem[Khan et~al.(2024)Khan, Xu, Liu, Lagstedt, Alam{\"a}ki, and Kauttonen]{khan2024exploring}
Umair~Ali Khan, Qianru Xu, Yang Liu, Altti Lagstedt, Ari Alam{\"a}ki, and Janne Kauttonen.
\newblock Exploring contactless techniques in multimodal emotion recognition: insights into diverse applications, challenges, solutions, and prospects.
\newblock \emph{Multimedia Systems}, 30\penalty0 (3):\penalty0 115, 2024.

\bibitem[Khattak et~al.(2023)Khattak, Rasheed, Maaz, Khan, and Khan]{khattak2023maple}
Muhammad~Uzair Khattak, Hanoona Rasheed, Muhammad Maaz, Salman Khan, and Fahad~Shahbaz Khan.
\newblock Maple: Multi-modal prompt learning.
\newblock In \emph{Proceedings of the IEEE/CVF Conference on Computer Vision and Pattern Recognition}, pages 19113--19122, 2023.

\bibitem[Kiela et~al.(2020)Kiela, Firooz, Mohan, Goswami, Singh, Ringshia, and Testuggine]{kiela2020hateful}
Douwe Kiela, Hamed Firooz, Aravind Mohan, Vedanuj Goswami, Amanpreet Singh, Pratik Ringshia, and Davide Testuggine.
\newblock The hateful memes challenge: Detecting hate speech in multimodal memes.
\newblock \emph{Advances in neural information processing systems}, pages 2611--2624, 2020.

\bibitem[Kim and Kim(2024)]{kim2024missing}
Donggeun Kim and Taesup Kim.
\newblock Missing modality prediction for unpaired multimodal learning via joint embedding of unimodal models.
\newblock In \emph{European Conference on Computer Vision}, pages 171--187, 2024.

\bibitem[Kim et~al.(2024)Kim, Kim, Moon, Choi, and Kim]{kim2024you}
Minkuk Kim, Hyeon~Bae Kim, Jinyoung Moon, Jinwoo Choi, and Seong~Tae Kim.
\newblock Do you remember? dense video captioning with cross-modal memory retrieval.
\newblock In \emph{Proceedings of the IEEE/CVF Conference on Computer Vision and Pattern Recognition}, pages 13894--13904, 2024.

\bibitem[Lang et~al.(2025)Lang, Cheng, Zhong, and Zhou]{lang2025retrieval}
Jian Lang, Zhangtao Cheng, Ting Zhong, and Fan Zhou.
\newblock Retrieval-augmented dynamic prompt tuning for incomplete multimodal learning.
\newblock \emph{arXiv preprint arXiv:2501.01120}, 2025.

\bibitem[Le et~al.(2024)Le, Thwal, Qiao, Tun, Nguyen, and Hong]{le2024cross}
Huy~Q Le, Chu~Myaet Thwal, Yu Qiao, Ye~Lin Tun, Minh~NH Nguyen, and Choong~Seon Hong.
\newblock Cross-modal prototype based multimodal federated learning under severely missing modality.
\newblock \emph{arXiv preprint arXiv:2401.13898}, 2024.

\bibitem[Lee et~al.(2023)Lee, Tsai, Chiu, and Lee]{lee2023multimodal}
Yi-Lun Lee, Yi-Hsuan Tsai, Wei-Chen Chiu, and Chen-Yu Lee.
\newblock Multimodal prompting with missing modalities for visual recognition.
\newblock In \emph{Proceedings of the IEEE/CVF Conference on Computer Vision and Pattern Recognition}, pages 14943--14952, 2023.

\bibitem[Li et~al.(2024{\natexlab{a}})Li, Jin, Sun, Yu, Shi, Hao, Hao, Liu, Sun, Zhang, et~al.]{li2024foundation}
Dingzhe Li, Yixiang Jin, Yuhao Sun, Hongze Yu, Jun Shi, Xiaoshuai Hao, Peng Hao, Huaping Liu, Fuchun Sun, Jianwei Zhang, et~al.
\newblock What foundation models can bring for robot learning in manipulation: A survey.
\newblock \emph{arXiv preprint arXiv:2404.18201}, 2024{\natexlab{a}}.

\bibitem[Li et~al.(2024{\natexlab{b}})Li, Peng, Chen, Gao, and Yang]{li2024configure}
Li Li, Jiawei Peng, Huiyi Chen, Chongyang Gao, and Xu Yang.
\newblock How to configure good in-context sequence for visual question answering.
\newblock In \emph{Proceedings of the IEEE/CVF Conference on Computer Vision and Pattern Recognition}, pages 26710--26720, 2024{\natexlab{b}}.

\bibitem[Liu et~al.(2025)Liu, Hu, Li, Yi, Chang, Gao, and Yin]{liu2025tackling}
Tengfei Liu, Yongli Hu, Mingjie Li, Junfei Yi, Xiaojun Chang, Junbin Gao, and Baocai Yin.
\newblock Tackling real-world complexity: Hierarchical modeling and dynamic prompting for multimodal long document classification.
\newblock \emph{IEEE Transactions on Circuits and Systems for Video Technology}, 2025.

\bibitem[Liu et~al.(2023)Liu, Li, and Lin]{liu2023cross}
Yang Liu, Guanbin Li, and Liang Lin.
\newblock Cross-modal causal relational reasoning for event-level visual question answering.
\newblock \emph{IEEE Transactions on Pattern Analysis and Machine Intelligence}, 45\penalty0 (10):\penalty0 11624--11641, 2023.

\bibitem[Liu et~al.(2021)Liu, Lin, Cao, Hu, Wei, Zhang, Lin, and Guo]{liu2021swin}
Ze Liu, Yutong Lin, Yue Cao, Han Hu, Yixuan Wei, Zheng Zhang, Stephen Lin, and Baining Guo.
\newblock Swin transformer: Hierarchical vision transformer using shifted windows.
\newblock In \emph{Proceedings of the IEEE/CVF international conference on computer vision}, pages 10012--10022, 2021.

\bibitem[Loshchilov and Hutter(2017)]{loshchilov2017decoupled}
Ilya Loshchilov and Frank Hutter.
\newblock Decoupled weight decay regularization.
\newblock \emph{arXiv preprint arXiv:1711.05101}, 2017.

\bibitem[Meng et~al.(2024)Meng, Sun, Xu, He, and Shen]{meng2024multi}
Xiangxi Meng, Kaicong Sun, Jun Xu, Xuming He, and Dinggang Shen.
\newblock Multi-modal modality-masked diffusion network for brain mri synthesis with random modality missing.
\newblock \emph{IEEE Transactions on Medical Imaging}, 2024.

\bibitem[Pipoli et~al.(2025)Pipoli, Bolelli, Sarto, Cornia, Baraldi, Grana, Cucchiara, Ficarra, et~al.]{pipoli2025semantically}
Vittorio Pipoli, Federico Bolelli, Sara Sarto, Marcella Cornia, Lorenzo Baraldi, Costantino Grana, Rita Cucchiara, Elisa Ficarra, et~al.
\newblock Semantically conditioned prompts for visual recognition under missing modality scenarios.
\newblock In \emph{Proceedings of the IEEE/CVF Winter Conference on Applications of Computer Vision}, 2025.

\bibitem[Radford et~al.(2021)Radford, Kim, Hallacy, Ramesh, Goh, Agarwal, Sastry, Askell, Mishkin, Clark, et~al.]{radford2021learning}
Alec Radford, Jong~Wook Kim, Chris Hallacy, Aditya Ramesh, Gabriel Goh, Sandhini Agarwal, Girish Sastry, Amanda Askell, Pamela Mishkin, Jack Clark, et~al.
\newblock Learning transferable visual models from natural language supervision.
\newblock In \emph{International conference on machine learning}, pages 8748--8763, 2021.

\bibitem[Shi et~al.(2025)Shi, Feng, Shang, Wan, et~al.]{shi2025deep}
Tongkai Shi, Wei Feng, Fanhua Shang, Liang Wan, et~al.
\newblock Deep correlated prompting for visual recognition with missing modalities.
\newblock \emph{Advances in Neural Information Processing Systems}, pages 67446--67466, 2025.

\bibitem[Sima et~al.(2024)Sima, Renz, Chitta, Chen, Zhang, Xie, Bei{\ss}wenger, Luo, Geiger, and Li]{sima2024drivelm}
Chonghao Sima, Katrin Renz, Kashyap Chitta, Li Chen, Hanxue Zhang, Chengen Xie, Jens Bei{\ss}wenger, Ping Luo, Andreas Geiger, and Hongyang Li.
\newblock Drivelm: Driving with graph visual question answering.
\newblock In \emph{European Conference on Computer Vision}, pages 256--274, 2024.

\bibitem[Song et~al.(2024)Song, Guo, Yang, Tang, and Wang]{song2024emotional}
Peipei Song, Dan Guo, Xun Yang, Shengeng Tang, and Meng Wang.
\newblock Emotional video captioning with vision-based emotion interpretation network.
\newblock \emph{IEEE Transactions on Image Processing}, pages 1122--1135, 2024.

\bibitem[Sun et~al.(2025)Sun, Zheng, Li, Gao, Nie, Huang, and Wei]{sun2025strong}
Tianci Sun, Chengyu Zheng, Xiu Li, Yanli Gao, Jie Nie, Lei Huang, and Zhiqiang Wei.
\newblock Strong and weak prompt engineering for remote sensing image-text cross-modal retrieval.
\newblock \emph{IEEE Journal of Selected Topics in Applied Earth Observations and Remote Sensing}, 2025.

\bibitem[Tan et~al.(2025)Tan, Ji, Hao, Lin, Wang, Wang, and Zhang]{tan2025reason}
Huajie Tan, Yuheng Ji, Xiaoshuai Hao, Minglan Lin, Pengwei Wang, Zhongyuan Wang, and Shanghang Zhang.
\newblock Reason-rft: Reinforcement fine-tuning for visual reasoning.
\newblock \emph{arXiv preprint arXiv:2503.20752}, 2025.

\bibitem[Van~der Maaten and Hinton(2008)]{van2008visualizing}
Laurens Van~der Maaten and Geoffrey Hinton.
\newblock Visualizing data using t-sne.
\newblock \emph{Journal of machine learning research}, 9\penalty0 (11), 2008.

\bibitem[Wang et~al.(2015)Wang, Kumar, Thome, Cord, and Precioso]{wang2015recipe}
Xin Wang, Devinder Kumar, Nicolas Thome, Matthieu Cord, and Frederic Precioso.
\newblock Recipe recognition with large multimodal food dataset.
\newblock In \emph{IEEE International Conference on Multimedia \& Expo Workshops}, pages 1--6, 2015.

\bibitem[Wu et~al.(2024)Wu, Wang, Chen, and Carneiro]{wu2024deep}
Renjie Wu, Hu Wang, Hsiang-Ting Chen, and Gustavo Carneiro.
\newblock Deep multimodal learning with missing modality: A survey.
\newblock \emph{arXiv preprint arXiv:2409.07825}, 2024.

\bibitem[Yang et~al.(2025)Yang, Yin, Gu, Deng, Zhang, and Zhu]{yang2025consistent}
Muli Yang, Jie Yin, Yanan Gu, Cheng Deng, Hanwang Zhang, and Hongyuan Zhu.
\newblock Consistent prompt tuning for generalized category discovery.
\newblock \emph{International Journal of Computer Vision}, pages 1--28, 2025.

\bibitem[Yao et~al.(2024)Yao, Zhang, Zhang, Liu, Chua, and Sun]{yao2024cpt}
Yuan Yao, Ao Zhang, Zhengyan Zhang, Zhiyuan Liu, Tat-Seng Chua, and Maosong Sun.
\newblock Cpt: Colorful prompt tuning for pre-trained vision-language models.
\newblock \emph{AI Open}, 5:\penalty0 30--38, 2024.

\bibitem[Zhang et~al.(2024{\natexlab{a}})Zhang, Wu, Gao, Shen, and Song]{zhang2024dept}
Ji Zhang, Shihan Wu, Lianli Gao, Heng~Tao Shen, and Jingkuan Song.
\newblock Dept: Decoupled prompt tuning.
\newblock In \emph{Proceedings of the IEEE/CVF Conference on Computer Vision and Pattern Recognition}, pages 12924--12933, 2024{\natexlab{a}}.

\bibitem[Zhang et~al.(2024{\natexlab{b}})Zhang, Pan, Yang, Li, and Chen]{zhang2024unified}
Yupei Zhang, Li Pan, Qiushi Yang, Tan Li, and Zhen Chen.
\newblock Unified multi-modal diagnostic framework with reconstruction pre-training and heterogeneity-combat tuning.
\newblock \emph{IEEE Journal of Biomedical and Health Informatics}, 2024{\natexlab{b}}.

\bibitem[Zhao et~al.(2021)Zhao, Li, and Jin]{zhao2021missing}
Jinming Zhao, Ruichen Li, and Qin Jin.
\newblock Missing modality imagination network for emotion recognition with uncertain missing modalities.
\newblock In \emph{Proceedings of the 59th Annual Meeting of the Association for Computational Linguistics and the 11th International Joint Conference on Natural Language Processing (Volume 1: Long Papers)}, pages 2608--2618, 2021.

\bibitem[Zhao et~al.(2024)Zhao, Zhang, Ma, and Cheng]{zhao2024survey}
Tianyi Zhao, Liangliang Zhang, Yao Ma, and Lu Cheng.
\newblock A survey on safe multi-modal learning systems.
\newblock In \emph{Proceedings of the 30th ACM SIGKDD Conference on Knowledge Discovery and Data Mining}, pages 6655--6665, 2024.

\bibitem[Zhou et~al.(2022)Zhou, Yang, Loy, and Liu]{zhou2022learning}
Kaiyang Zhou, Jingkang Yang, Chen~Change Loy, and Ziwei Liu.
\newblock Learning to prompt for vision-language models.
\newblock \emph{International Journal of Computer Vision}, pages 2337--2348, 2022.

\bibitem[Zhou et~al.(2024)Zhou, Liang, Xu, Zhu, Xu, Zou, and Bai]{zhou2024dynamic}
Xin Zhou, Dingkang Liang, Wei Xu, Xingkui Zhu, Yihan Xu, Zhikang Zou, and Xiang Bai.
\newblock Dynamic adapter meets prompt tuning: Parameter-efficient transfer learning for point cloud analysis.
\newblock In \emph{Proceedings of the IEEE/CVF Conference on Computer Vision and Pattern Recognition}, pages 14707--14717, 2024.

\end{thebibliography}
}

\clearpage

\newpage

\appendix

\begin{table*}
 \renewcommand\arraystretch{1.3}
\centering
\resizebox{\textwidth}{!}{ 
\begin{tabular}{|l|l|l|l|l|l|}
\hline
\textbf{Dataset} & \textbf{Task} & \textbf{Modalities} & \textbf{Samples} & \textbf{Train/Val/Test Split} & \textbf{Key Features} \\ \hline
\textbf{MM-IMDb}~\cite{arevalo2017gated} & Movie genre classification & Image, Text & 25,959 movies & 15,552 / 2,608 / 7,799 & Multi-label classification; each movie has a poster image and textual metadata. \\ \hline
\textbf{Hateful Memes}~\cite{kiela2020hateful} & Hateful meme detection & Image, Text & 10,000 memes & 8,500 / 500 / 1,500 & Challenges unimodal models; text and image must be jointly analyzed. \\ \hline
\textbf{UPMC Food101}~\cite{wang2015recipe} & Food classification & Image, Text & 90,688 image-text pairs & 67,972 / - / 22,716 & Aligns with ETHZ Food-101; noisy image-text pairs from Google Image Search. \\ \hline
\end{tabular}
}
\caption{Overview of the multimodal datasets used in this work.}
\label{tab:datasets}
\end{table*}

\begin{figure*}
    \centering
    \includegraphics[width=0.92\linewidth]{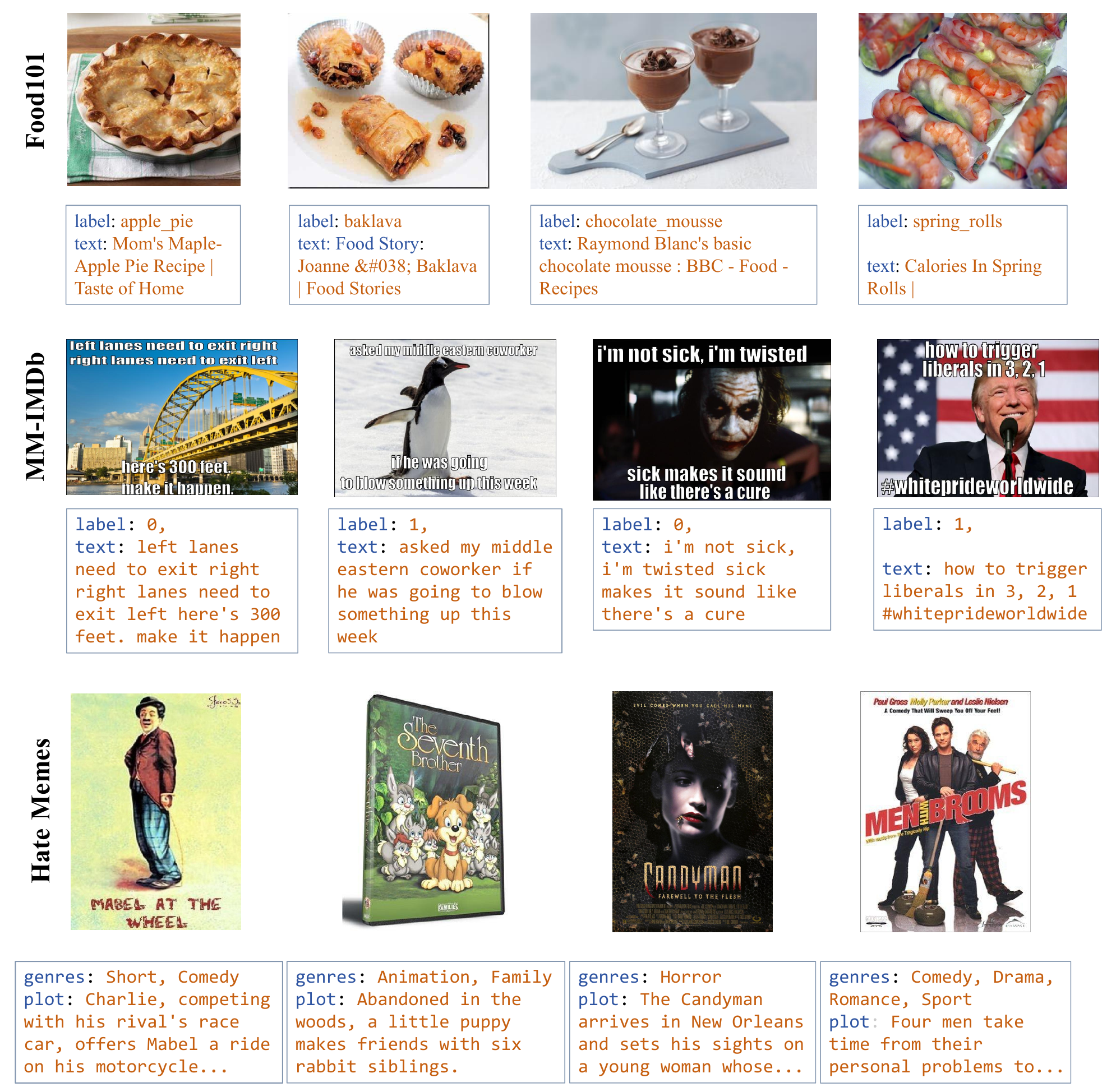}
    \caption{Visualization of the samples in three dataset.}
    \label{fig:date_hate}
\end{figure*}

This supplementary material provides additional details on the proposed SyP and experimental results that could not be included in the main manuscript due to page limitations.
Specifically, this appendix is organized as follows:

\begin{itemize}
\item Sec.~\ref{sec1} presents details of the training datasets.
\item Sec.~\ref{sec2} presents additional details of training strategies.
\item Sec.~\ref{sec3} complements more experiment results and analysis.
\item Sec.~\ref{sec4} shows more visualization results to prove the effectiveness of \textit{\textbf{SyP}}.
\end{itemize}

\section{Details of Training Datasets}
\label{sec1}

To evaluate the effectiveness of the proposed SyP, we conduct experiments on three widely-used multimodal datasets: \textbf{MM-IMDb}, \textbf{Hateful Memes}, and \textbf{UPMC Food-101}. These datasets represent diverse tasks, including movie genre classification, hateful meme detection, and food classification, and are designed to challenge models in handling multimodal data under varying conditions. Below, we provide an overview of each dataset, highlighting their key characteristics, modalities, and task-specific challenges. 
A summary is presented in Tab.~\ref{tab:datasets}, and visualization of the samples in the three datasets is shown in Fig.~\ref{fig:date_hate}.

\textbf{MM-IMDb} is the largest publicly available dataset for movie genre classification, featuring 25,959 movies annotated with both poster images and textual metadata. It is a multi-label classification task, as movies can belong to multiple genres simultaneously. The dataset is split into 15,552 training, 2,608 validation, and 7,799 test samples.

\textbf{Hateful Memes}, curated by Facebook AI, is designed for hateful meme detection. It contains 10,000 multimodal examples where both text and image contribute to the overall meaning. The dataset is structured to challenge unimodal models, with 8,500 samples for training, 500 for validation, and 1,500 for testing.

\textbf{UPMC Food-101} is a large-scale food classification dataset comprising 90,688 noisy image-text pairs collected from Google Image Search. It spans 101 food categories and aligns with the ETHZ Food-101 dataset. The dataset includes 67,972 training samples and 22,716 test samples, with no designated validation set. Each sample consists of an image and an accompanying textual description.

\section{Training Details}
\label{sec2}

All experiments are performed on four NVIDIA RTX 3090 GPUs. We employ the AdamW optimizer~\cite{loshchilov2017decoupled} with an initial learning rate of $1e^{-3}$ and a weight decay of $2e^{-2}$. The learning rate is warmed up for 10\% of the total training steps and then decays linearly to zero. We set the prompt depth as 36. Experiments are conducted on three datasets: Hateful Memes, Food101, and MM-IMDb, using a frozen backbone strategy to efficiently fine-tune CLIP for multimodal tasks while preserving knowledge from large-scale vision-language pretraining.

\begin{itemize}
    \item \textbf{Hateful Memes}: The dataset is trained with a batch size of 256 for 20 epochs (10,000 steps). The validation check interval is 11\%, and the optimizer uses a learning rate of 1e-2. The maximum text length is 128.
    
    \item \textbf{Food101}: The dataset is trained with a batch size of 256 for 200 epochs (20,000 steps). The validation check interval is 20\%, and the optimizer uses a learning rate of 1e-2. The maximum text length is 512.
    
    \item \textbf{MM-IMDb}: The dataset is trained with a batch size of 256 for 100 epochs (25,000 steps). The validation check interval is 20\%, and the optimizer uses a learning rate of 1e-4 with a weight decay of 0.01. The maximum text length is 40, and a prompt-based method is used to handle missing modalities. The image encoder is ViT-B/16, with a vocabulary size of 30,522. Whole word masking is disabled, the masked language modeling probability is set to 15\%, and both training and validation image transformations use CLIP\_transform.
\end{itemize}

This setup ensures efficient fine-tuning of CLIP for diverse multimodal tasks while leveraging the benefits of large-scale pretraining. The frozen backbone strategy reduces computational costs and maintains the model's generalization capabilities.

\begin{figure*}\small
    \centering
    \includegraphics[width=1\linewidth]{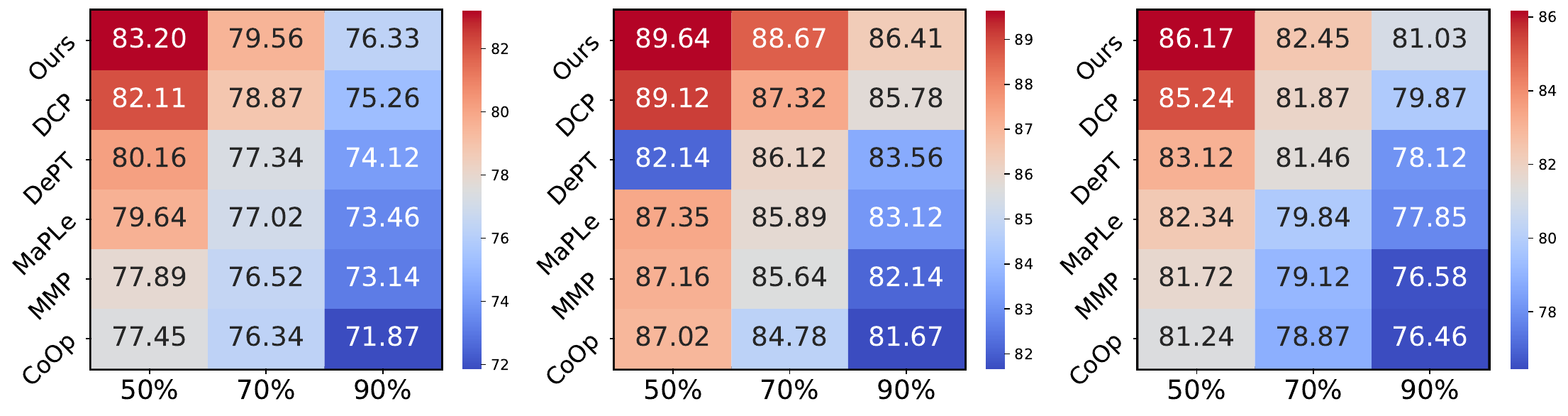}
    (a) Text Missing \hspace{3.5cm} (b) Image Missing \hspace{3.5cm} (c) Both Missing
    \caption{Generalization analysis on the Food101 dataset across various missing rates in terms of Accuracy.}
    \label{fig:heatmap2}
\end{figure*}

\begin{figure*}\small
    \centering
    \includegraphics[width=1\linewidth]{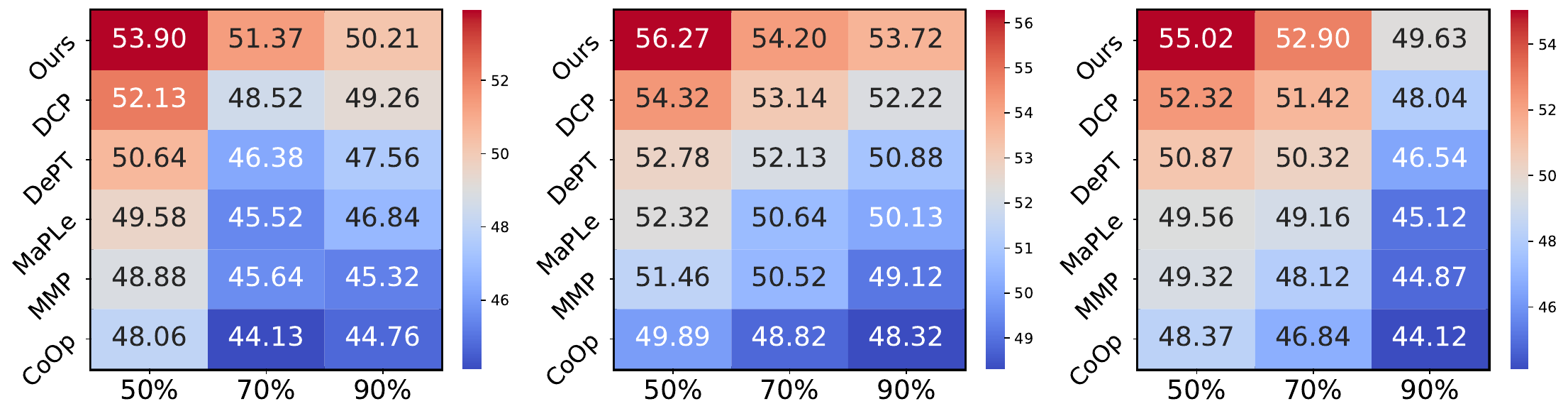}
    (a) Text Missing \hspace{3.5cm} (b) Image Missing \hspace{3.5cm} (c) Both Missing
    \caption{Generalization analysis on the MM-IMDb dataset across various missing rates in terms of F1\_Macro.}
    \label{fig:heatmap3}
\end{figure*}

\section{Additional Experiments and Analysis}
\label{sec3}

\textbf{Model Generalizability}
We evaluate the model's generalizability on Food101 and MM-IMDb datasets under varying missing modality rates (50\%, 70\%, 90\%). 
Results are shown in Fig.~\ref{fig:heatmap2} (Accuracy for Food101) and Fig.~\ref{fig:heatmap3} (F1\_Macro for MM-IMDb).

For the Food101 dataset (Fig.~\ref{fig:heatmap2}), the model demonstrates strong performance across all missing conditions. In text-missing scenarios, it maintains high accuracy by effectively leveraging image features, even at a 90\% missing rate. Similarly, in image-missing cases, the model relies on textual data to sustain robust performance. Even when both modalities are missing, the model adapts well, inferring missing information from available data, showcasing its dynamic adaptation capabilities.
For the MM-IMDb dataset (Fig.~\ref{fig:heatmap3}), the model exhibits consistent performance despite missing data. In text-missing conditions, it effectively uses visual cues, while in image-missing scenarios, it compensates with textual features. In the most challenging case of both modalities missing, the model still performs competitively by intelligently utilizing remaining data, highlighting its resilience.

These results underscore the model's adaptability to incomplete multimodal data, making it suitable for real-world applications where data completeness is often uncertain. 
The consistent performance across datasets and conditions validates the effectiveness of its dynamic adapter and synergistic prompts strategy.

\begin{figure*}\small
    \centering
    \includegraphics[width=0.98\linewidth]{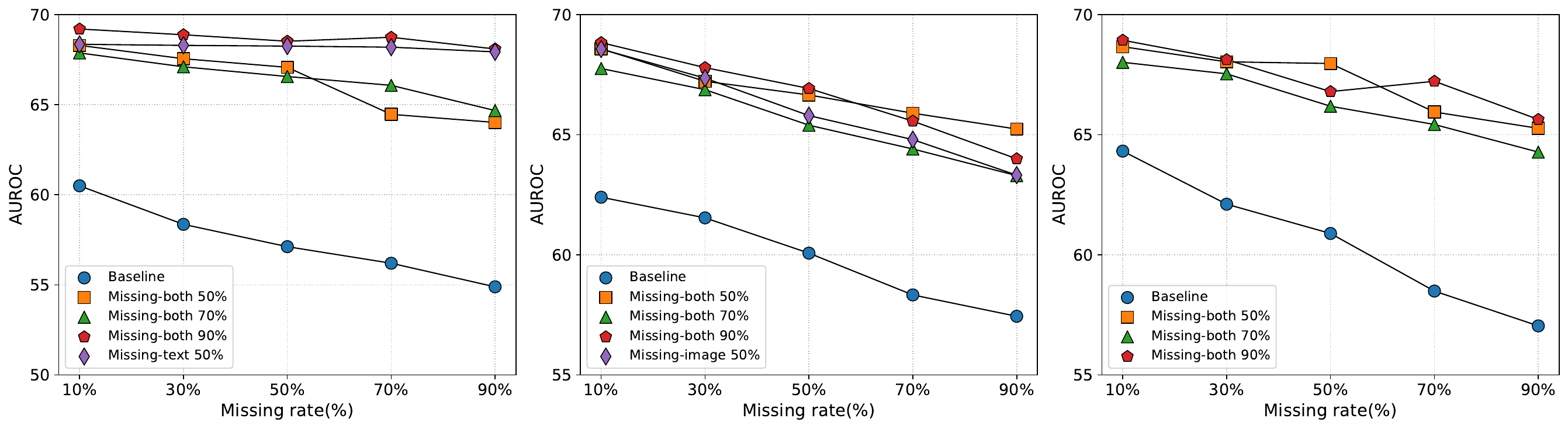}
    (a) Text Missing \hspace{3.5cm} (b) Image Missing \hspace{3.5cm} (c) Both Missing
    \caption{\textbf{Generalizability Analysis of Our Method to Different Missing Rates on Hate Memes dataset.} (a) Models are trained on missing-both or missing-text cases, and evaluated on missing-text cases with different missing rates. (b) Models are trained on missing-both or missing-image cases, and evaluated on missing-image cases with different missing rates. (c) All models are trained on missing-both cases, and evaluated on missing-both cases with different missing rates.}
    \label{fig:line_chart_hate}
\end{figure*}

\begin{figure*}\small
    \centering
    \includegraphics[width=0.98\linewidth]{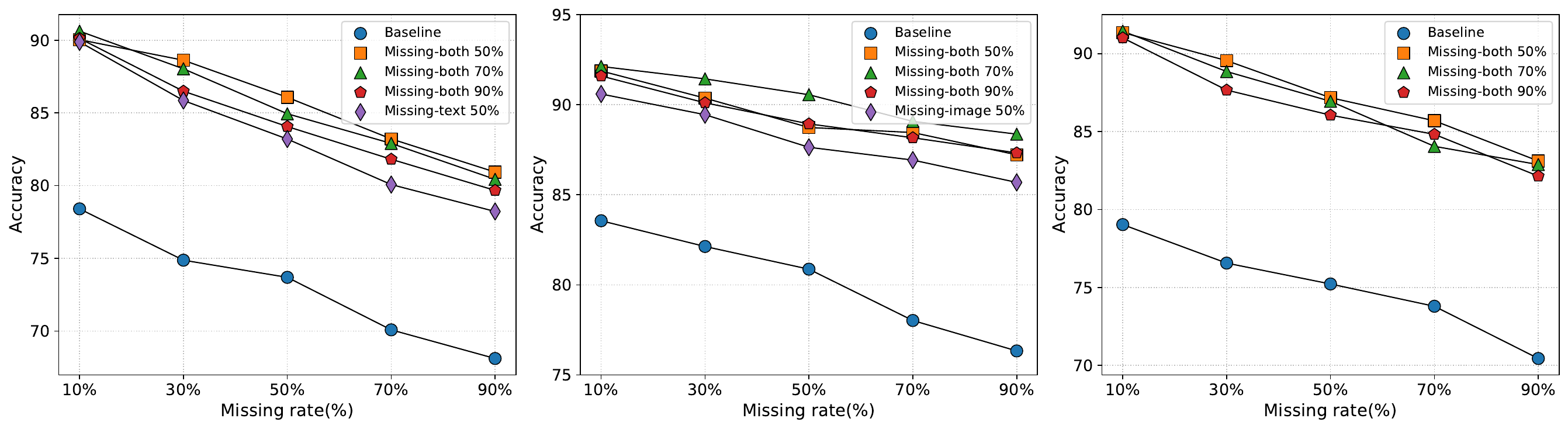}
    (a) Text Missing \hspace{3.5cm} (b) Image Missing \hspace{3.5cm} (c) Both Missing
    \caption{\textbf{Generalizability Analysis of Our Method to Different Missing Rates on Food101 dataset.} (a) Models are trained on missing-both or missing-text cases, and evaluated on missing-text cases with different missing rates. (b) Models are trained on missing-both or missing-image cases, and evaluated on missing-image cases with different missing rates. (c) All models are trained on missing-both cases, and evaluated on missing-both cases with different missing rates.}
    \label{fig:line_chart_food}
\end{figure*}

\noindent \textbf{Generalizability to Different Missing Rates} 
To validate the generalizability of the proposed SyP across different missing rates, we conduct experiments on two distinct multi-modal datasets: the \textit{Hateful Memes} dataset and the \textit{Food101} dataset. The experimental results, presented in Figs.~\ref{fig:line_chart_hate} and~\ref{fig:line_chart_food}, provide valuable insights into the performance of SyP across various missing-modality scenarios, including text missing, image missing, and both missing.

On the \textit{Hateful Memes} dataset, as depicted in Fig.~\ref{fig:line_chart_hate}, SyP variants consistently outperform baseline models across all missing rates (10\%--90\%). This significant improvement underscores the robustness of the proposed SyP in handling incomplete data. 
Notably, models trained both on a single missing modality and under missing-both conditions demonstrate strong and stable performance across all scenarios, especially in the text missing scenarios. This indicates that the combination of static and dynamic prompts enables flexible adaptation, particularly when trained with higher missing rates.
Similarly, on the \textit{Food101} dataset, SyP demonstrates remarkable performance. As shown in Fig.~\ref{fig:line_chart_food}, SyP achieves the highest accuracy across all missing rates and scenarios. The dynamic adapter effectively adjusts the prompt weights, allowing the model to handle diverse missing-modality situations with high accuracy. These results highlight the adaptability and effectiveness of SyP in real-world applications where data may be noisy or incomplete.

\begin{figure*}
    \centering
    \includegraphics[width=0.98\linewidth]{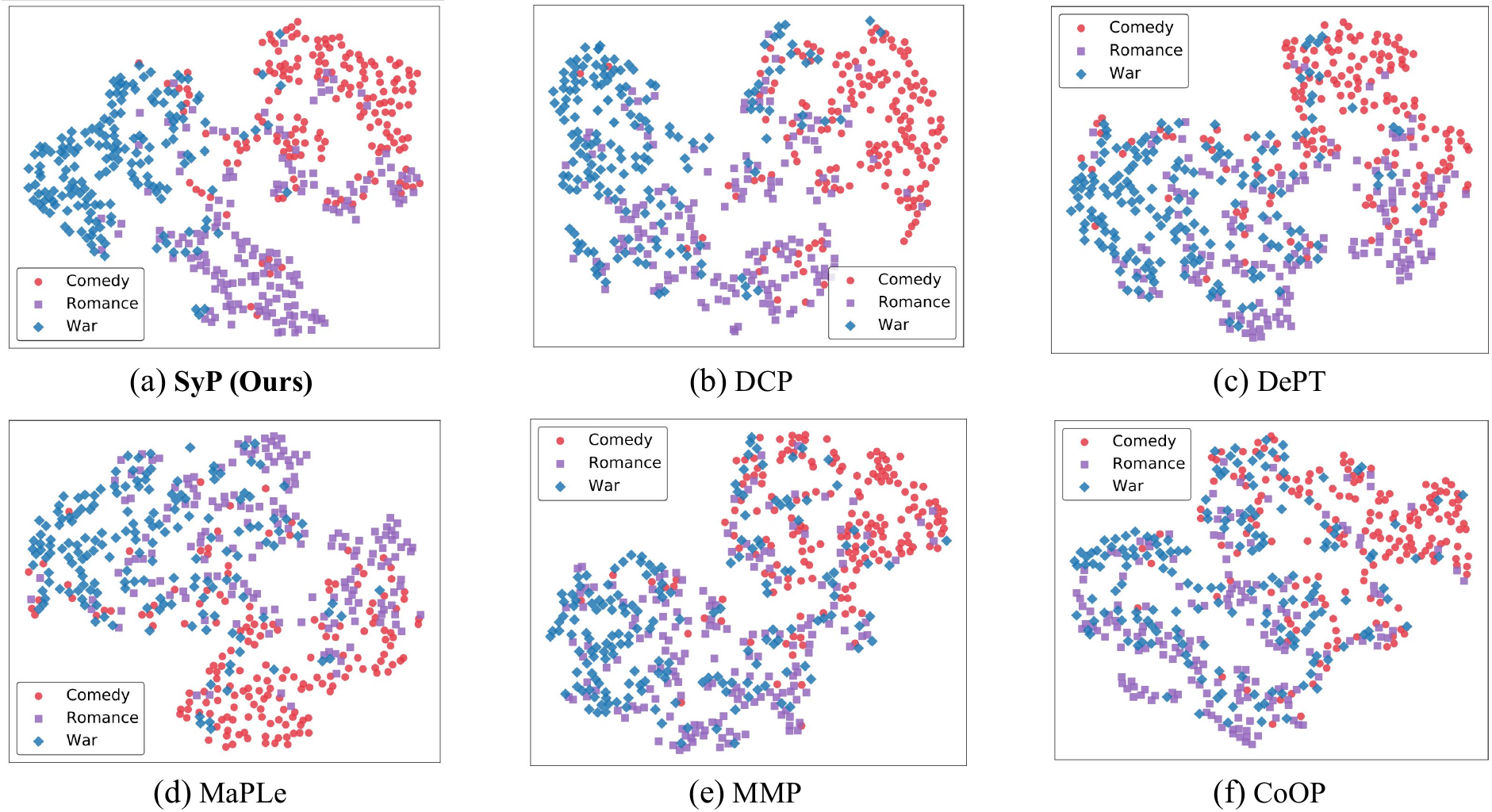}
    \caption{t-SNE visualization of our model and other baselines on the MM-IMDb dataset under a 50\% both missing rate.}
    \label{fig:TSNE}
\end{figure*}

\begin{table}
    \centering
    \setlength{\tabcolsep}{2.5pt}
    \begin{tabular}{cccc}
    \toprule
       {\textbf{Reduction}}   & \textbf{Hateful Memes} & \textbf{Food101} & \textbf{MM-IMDb} \\
        \textbf{Ratio} &\textbf{(AUROC)} & \textbf{(Accuracy)} & \textbf{(F1\_Macro)}\\
       \midrule
        \(r\) = 5 & 67.98 & \textbf{86.17} & 54.72\\
        \(r\) = 10 & 67.69 & 85.83 &\textbf{55.02} \\
        \(r\) = 16 & \textbf{68.16} & 85.58 & 53.98\\
        \bottomrule
        
    \end{tabular}
    \caption{Hyper-parameter analysis of reduction ratio \(r\) under a 50\% both missing case on three datasets.}
    \vspace{-1em}
    \label{tab:ablation2}
\end{table}

\noindent \textbf{Hyper-Parameter Analysis}
We analyze the impact of the reduction ratio \(r\) in the dynamic adapter under a 50\% missing-modality scenario across three datasets: Hateful Memes, Food101, and MM-IMDb. As shown in Tab.~\ref{tab:ablation2}, the reduction ratio \(r\) significantly influences model performance. For Hateful Memes, a ratio of 16 achieves the highest AUROC (68.16), indicating that larger ratios better capture subtle visual features. Food101 performs best with a ratio of 5 (86.17\% accuracy), suggesting smaller ratios are more effective for fine-grained classification. For MM-IMDb, a ratio of 10 yields the highest F1-Macro (55.02), demonstrating that a balanced approach is optimal for text-heavy datasets. 
These results underscore the importance of tuning the reduction ratio to adapt the dynamic adapter to diverse task requirements, ensuring robust performance across varying data characteristics and modalities.

\section{Visualization}
\label{sec4}
Fig.~\ref{fig:TSNE} illustrates the t-SNE~\cite{van2008visualizing} visualization of the embedding distributions for three genres (Comedy, Romance, and War) in the MM-IMDb test set under a 50\% both missing rate.
The points corresponding to \textit{\textbf{SyP}} are more tightly clustered and clearly separated than those of DCP~\cite{shi2025deep}, DePT~\cite{zhang2024dept},
MaPLe~\cite{khattak2023maple},
MMP~\cite{lee2023multimodal},
and CoOp~\cite{zhou2022learning}. This indicates that the proposed SyP can effectively manage multi-modal missing problems. The model can accurately understand and reason when facing different modalities, maintaining high performance and stability.
Specifically, the t-SNE visualization of \textit{\textbf{SyP}} has several characteristics. First, the high concentration of points indicates the model's accurate recognition of similar samples. Second, the clear separation between different classes reduces misclassification. Third, the uniform distribution of points shows that the proposed
SyP can balance the relationships between classes, leading to more stable model outputs.
This demonstrates that the proposed SyP is better at maintaining robust genre distinctions, despite the challenge posed by missing data, showcasing its strength in dynamic multi-modal adaptation.

\end{document}